\documentclass[9pt,twocolumn]{article}
\usepackage[a4paper, margin=1in]{geometry}
\usepackage{sectsty}
\sectionfont{\normalsize}
\subsectionfont{\normalsize}
\subsubsectionfont{\normalsize}

\skip\footins 19.5pt plus 12pt minus 1pt

\usepackage[T1]{fontenc}
\usepackage{dfadobe}  
\usepackage{cite}  

\usepackage{xcolor}
\usepackage{xspace}
\usepackage{amsmath,amssymb}
\usepackage{ntheorem}
\usepackage{mathtools}
\usepackage{esint}
\usepackage{subfiles}
\usepackage{array}
\usepackage{booktabs}
\usepackage{hyperref}
\usepackage{cleveref} 
\usepackage{multirow}
\usepackage{wrapfig}
\usepackage{theorem}
\usepackage{algorithm}
\usepackage{algpseudocode}
\usepackage{import}
\usepackage{tikz}
\usepackage{float}
\usepackage{setspace}
\usepackage{placeins}
\usepackage{enumitem}
\usepackage{cuted}
\usepackage{caption}
\usepackage{authblk}
\usepackage[symbol]{footmisc}
\usepackage{wrapfig}

\newcommand{\jumpset}{J_{\nabla\phi}}
\newcommand{\jumpsetTheta}{J_{\nabla\phi_\theta}}
\newcommand{\R}{\mathbb{R}}
\newcommand{\N}{\mathbb{N}}
\newcommand{\surface}{\mathcal{S}}
\DeclareRobustCommand\onedot{\futurelet\@let@token\@onedot}

\newcommand{\eps}{\varepsilon}
\renewcommand{\d}{\textrm{d}}
\newtheorem{theorem}{Theorem}[section]

\title{Medial Axis Aware Learning of Signed Distance Functions}
\author[1]{Samuel Weidemaier\footnote{weidemaier@uni-bonn.de}}	
\author[1]{Christoph Norden-Smoch}
\author[1]{Martin Rumpf}

\affil[1]{Institute for Numerical Simulation, University of Bonn \vspace{-1.4cm}}

\date{}

\begin{document}

\maketitle
\begin{strip}	
	\centering
	\begin{tikzpicture}
	\node[inner sep=0pt] at (0.0, 0.0) {\includegraphics[width=0.9\linewidth, trim={0 1.2cm 0 0cm}, clip]{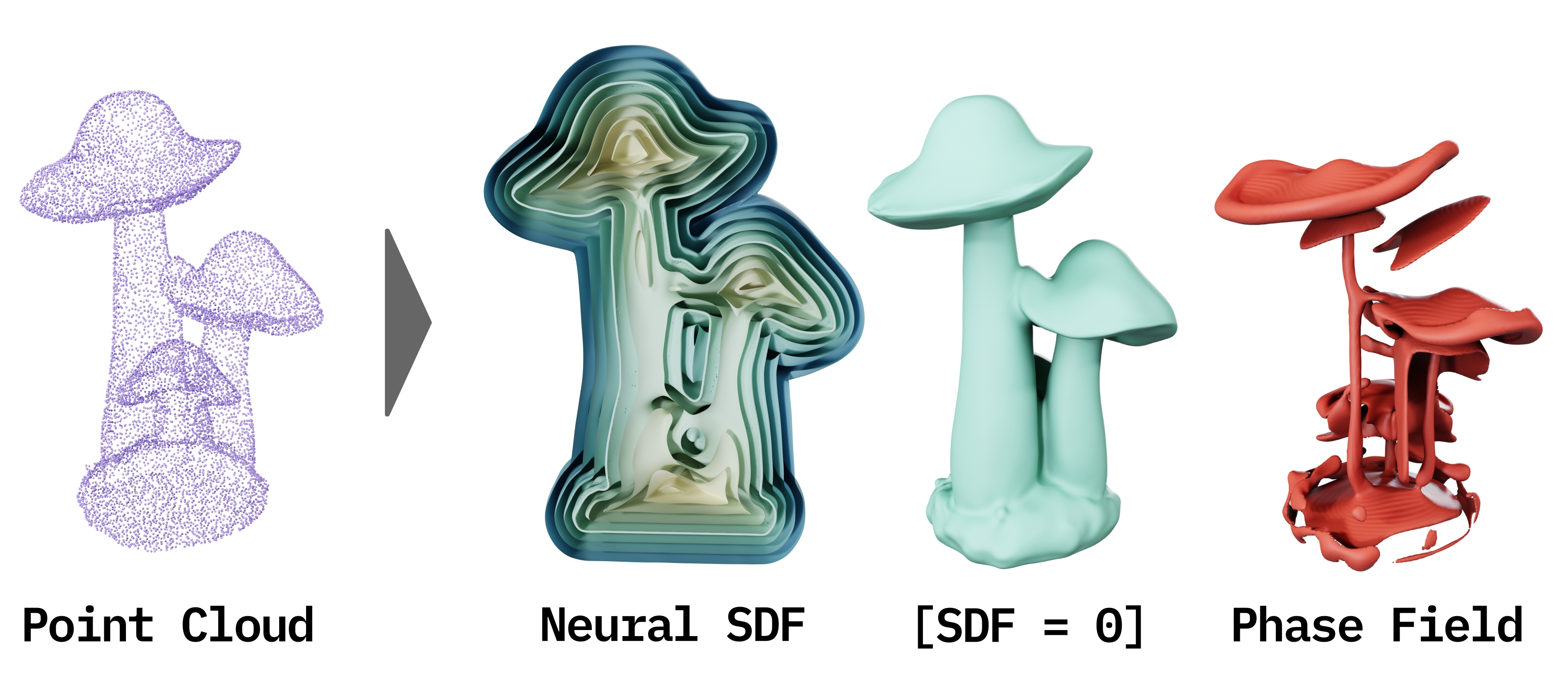}};
	\node[inner sep=0pt] at (-6.3, -3.4) {\textbf{Point cloud}};
	\node[inner sep=0pt] at (-1.1, -3.4) {\textbf{Neural SDF}};
	\node[inner sep=0pt] at (2.4, -3.4) {\textbf{[SDF = 0]}};
	\node[inner sep=0pt] at (6.2, -3.4) {\textbf{Phase field}};
	\end{tikzpicture}
	\captionof{figure}{We compute a global neural network approximation of the signed distance function (SDF) by taking into account a simultaneously trained neural network phase field representing the medial axis, i.e.
		the SDF gradient discontinuities.}
	\label{fig:teaser}
%
%
%
	\begin{abstract}\textit{
		We propose a novel variational method to compute a highly accurate 
		global signed distance function (SDF) to a given point cloud. 
		To this end, the jump set of the gradient of the SDF, which coincides with the medial axis of the surface, is explicitly taken into account through a higher-order variational formulation that enforces linear growth along the gradient direction away from this discontinuity set.
		The eikonal equation and the zero-level set of the SDF are enforced as constraints.
		To make this variational problem computationally tractable, 
		a phase field approximation of Ambrosio-Tortorelli type is employed. The associated phase field function implicitly describes the medial axis. 
		The method is implemented for surfaces represented by unoriented point clouds using neural network approximations of both the SDF and the phase field. 
		Experiments demonstrate the method's accuracy both in the near field and globally. 
		Quantitative and qualitative comparisons with other approaches show the advantages of the proposed method.}
	\end{abstract}
\end{strip}

\section{Introduction}
Computing the signed distance function (SDF) to an unoriented point cloud by training a neural network is a challenging task that has been addressed frequently in recent years, cf.~\cite{SIREN,Li21Phase,ben2022digs,yang2023steik,wang2023neural,hessian2,coiffier20241,HotSpot,HeatSDF26}.
It involves designing loss functionals such that the underlying surface can be represented accurately by the zero-level set, and at the same time, estimating the (signed) distance to the surface well. 

Applications of SDFs range from collision detection~\cite{liu2024real} and constructive solid geometry~\cite{marschner2023constructive} to the use of \emph{level set methods} for solving partial differential equations (PDEs) on surfaces~\cite{mehta2022level,HeatSDF26}. 
Furthermore, SDFs enable efficient ray tracing and rendering of implicit surfaces~\cite{SphereTracing}.

\begin{figure}[htbp!]
	\includegraphics[width=\linewidth]{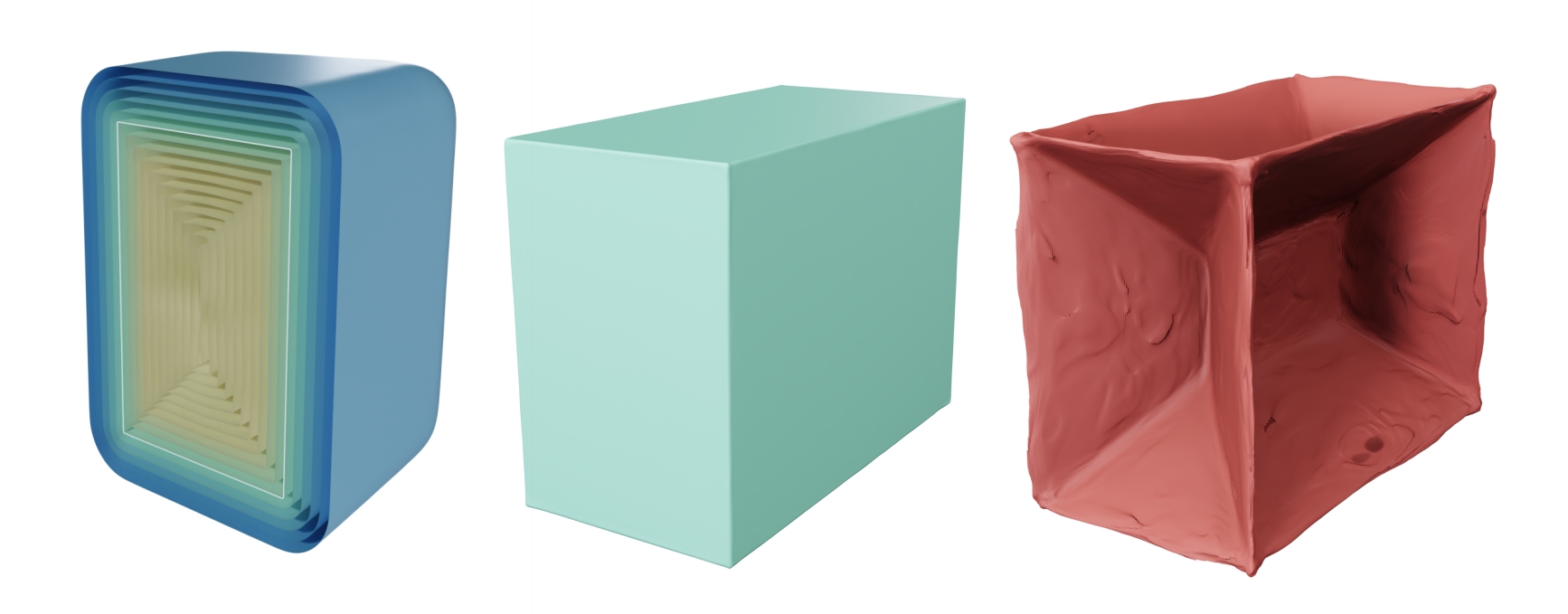}
	\caption{Left: level sets of the neural SDF $\phi$ of a hexahedron. Middle: zero-level set of the SDF. Right: phase field approximation of the jump set of the SDF gradient, which coincides with the hexaeder's medial axis.}
	\label{fig:Box_example}
\end{figure}

It is well known that the SDF is the \emph{viscosity} solution to the eikonal equation, a first-order hyperbolic PDE, which makes it difficult to design appropriate loss functionals.
The structure of the eikonal equation implies that solutions grow linearly in direction normal to the surface. Already for smooth closed surfaces, this leads to the fact that there exists a set, where the gradient of the SDF has to jump (cf.~\cref{fig:Box_example}). This set is the \emph{medial axis} of the surface, i.e., the lower-dimensional set, where the nearest-point projection onto the surface is not well-defined. Hence, a global regularization with a higher-order loss functional, including the medial axis, is not possible.  

{In addition to an eikonal loss, we introduce a second-order loss functional that encourages the approximate neural SDF to grow linearly in normal direction, together with a neural phase field function that detects the medial axis, where the gradient of the SDF jumps, and switches off the regularizer in those regions (cf. \cref{fig:Box_example}). This phase field is controlled by a loss functional of Ambrosio-Tortorelli-type, known in the context of phase field approximations of the Mumford-Shah functional for image segmentation. 
Our method computes a global neural network approximation of the signed distance function that is accurate both in the near field, allowing for high fidelity surface reconstruction, and in the far field.
Experimental results demonstrate that our approach achieves competitive or superior performance compared to state-of-the-art methods across both local and global metrics.}
\vspace{-0.3cm}
\subsection{Related work}

\paragraph*{Neural implicit representations.}
Implicit representations are a flexible tool in geometry processing \cite{essakine2025where}. Especially the implicit representation by the signed distance function is useful, in particular for the applications described above.
The method SIREN~\cite{SIREN} uses neural networks with sine activation function, together with an eikonal loss, to compute the SDF of a surface given by a point cloud. 
More recently, 1-Lipschitz neural implicit representations have been proposed to approximate signed distance functions while enforcing a global Lipschitz constraint by construction~\cite{coiffier20241}. This guarantees that the predicted distance cannot be overestimated and improves robustness of geometric queries even far from the surface.
In~\cite{Li21Phase}, the PHASE method is introduced to compute a neural implicit representation of a surface using an occupancy function, by minimizing the Modica-Mortola energy~\cite{MoMo77}. 
The resulting phase field function, represented by a neural network, exhibits two distinct phases corresponding to the interior and exterior regions of the surface. The phase transition occurs in a narrow band around the surface defined by the input point cloud.
In a post-processing step, the phase field function can be used to approximate the SDF, using a $\log$-transform.
\vspace{-0.1cm}
\paragraph*{Neural higher order variational SDF methods.}
The method DiGS~\cite{ben2022digs} modified the SIREN \cite{SIREN} approach to also include a second-order derivative loss functional, namely the squared $L^2$-norm of the Laplace operator. 
The singular structure of the Hessian of the SDF, induced by the linear growth in normal direction, was used in \cite{wang2023neural} and \cite{hessian2}, by introducing quadratic loss functionals depending on the determinant, and the second derivative in normal direction, respectively, in the near field of the surface. StEik~\cite{yang2023steik} also made use of this fact and introduced a second-order $L^1$-loss for the directional divergence in normal direction.
\vspace{-0.1cm}
\paragraph*{Heat methods for computing SDFs.}
Another approach for computing approximate distances, introduced by Crane et al.~\cite{crane2013geodesics,crane2017heat}, leverages the heat equation for geometry processing. This is based on a two-step method, by first solving the heat equation for a small time step, with the target object as the heat source. The gradients of the corresponding heat solution are then used to compute an approximate distance in a second step. Based on this approach, a method for computing parallel transport of tangent vectors on curved surfaces was presented in \cite{sharp2019vector}.
In~\cite{GSD}, the two-step method was generalized to compute SDFs, starting from an oriented point cloud. 
HeatSDF~\cite{HeatSDF26} adapts this two-step method in the learning context to compute an approximate neural SDF from \emph{unoriented} point clouds, incorporating additional constraints in order to orient the gradients. The HotSpot method~\cite{HotSpot} also uses the relation between the heat equation and distances in a one-step algorithm, by introducing an additional loss together with an eikonal loss, such that the SDF is encouraged to be the logarithm of the solution to the heat equation. 
\vspace{-0.1cm}
\paragraph*{Medial axis computation.}
The medial axis, also referred to as the \textit{skeleton}, has long been studied as a compact representation of shape geometry; see, e.g.,~\cite{Bl67}. More recently, neural approaches have been proposed to extract medial axes from different forms of input data, including point clouds and implicit representations, cf. ~\cite{PointSkeleton,ImplicitSkeleton}. In~\cite{NeuralSkeleton}, the authors introduce a effective algorithm for extracting a skeletal mesh from an implicit surface by iteratively following the gradient direction of the SDF from the surface until an singular point is reached. To this end, they use their own neural SDF method, which relies on a loss term involving the total variation of the gradient norm. 

\paragraph*{Higher order phase transition theory.}
In 1987, Aviles and Giga \cite{AvGi87} proposed a functional of Modica-Mortola-type~\cite{MoMo77}, with higher order derivatives,
in the context of smectic liquid crystal theory and conjectured the sharp-interface limit to be a functional that depends on the jump set of the gradient of a function that solves the eikonal equation almost everywhere. In two dimensions, the appropriate function space for the limit energy was investigated in \cite{AmDeMa99}, and the asymptotic sharp lower bound was studied in \cite{JinKohn2000}. In \cite{DeLellis02}, an example was given that the theory is fundamentally different in higher space dimensions. In \cite{AvGi96}, a connection of the minimizer of a sharp-interface limit energy to the viscosity solution of the eikonal equation was drawn.     

\paragraph*{Ambrosio-Tortorelli phase field models.}
The Mumford-Shah functional, cf. \cite{MuSh89}, is well-known in the context of image segmentation, allowing jumps of the image segmentation function along interfaces, and controlling the co-dimension one measure of the jump set. 
The phase field functional introduced by Ambrosio and Tortorelli, cf. \cite{AmTo92}, allows to approximate minimizers of the Mumford-Shah functional by smooth functions, where the jump set is approximated by a phase field function. 
Here, the term phase field for the smooth function $v$ emphasizes the 
preferred single phase $[v\approx1]$ in the bulk and the dropping down on a thin set.
In \cite{FrMa98}, a similar phase field model was introduced to approximate brittle fracture in the context of elasticity. Numerical approximation by finite elements was first discussed in \cite{BoFrMa00}. Recently, a discretization of a phase field fracture model based on a deep-Ritz method for neural networks was proposed in \cite{MaMoMi24}.
\vspace{-0.3cm}
\subsection{Our contribution}
\begin{itemize}
\item We introduce a variational approach to compute the global signed distance function of a surface represented by an unoriented point cloud and the discontinuity set of its gradient, i.e., the medial axis of the surface (cf.~\cref{fig:Box_example}).
The cost functional measures the  second-order derivative in normal direction where the SDF is smooth. Furthermore, it controls the $(d-1)$-dimensional measure of the discontinuity set.
\item We use an Ambrosio-Tortorelli type phase field model to variationally describe the medial axis together with a proper weighting of the second-order term. 
\item The method is implemented using neural networks for both the SDF and the phase field, which are trained simultaneously, allowing for an efficient computation.
\item We perform a quantitative evaluation and a comparison with other approaches for 2D and 3D databases, showing the high approximation quality of the SDF.
\end{itemize}
\vspace{-0.5cm}
\section{Motivation}\label{sec:motivation}
\begin{figure}[htbp!] 
\includegraphics[width=\linewidth]{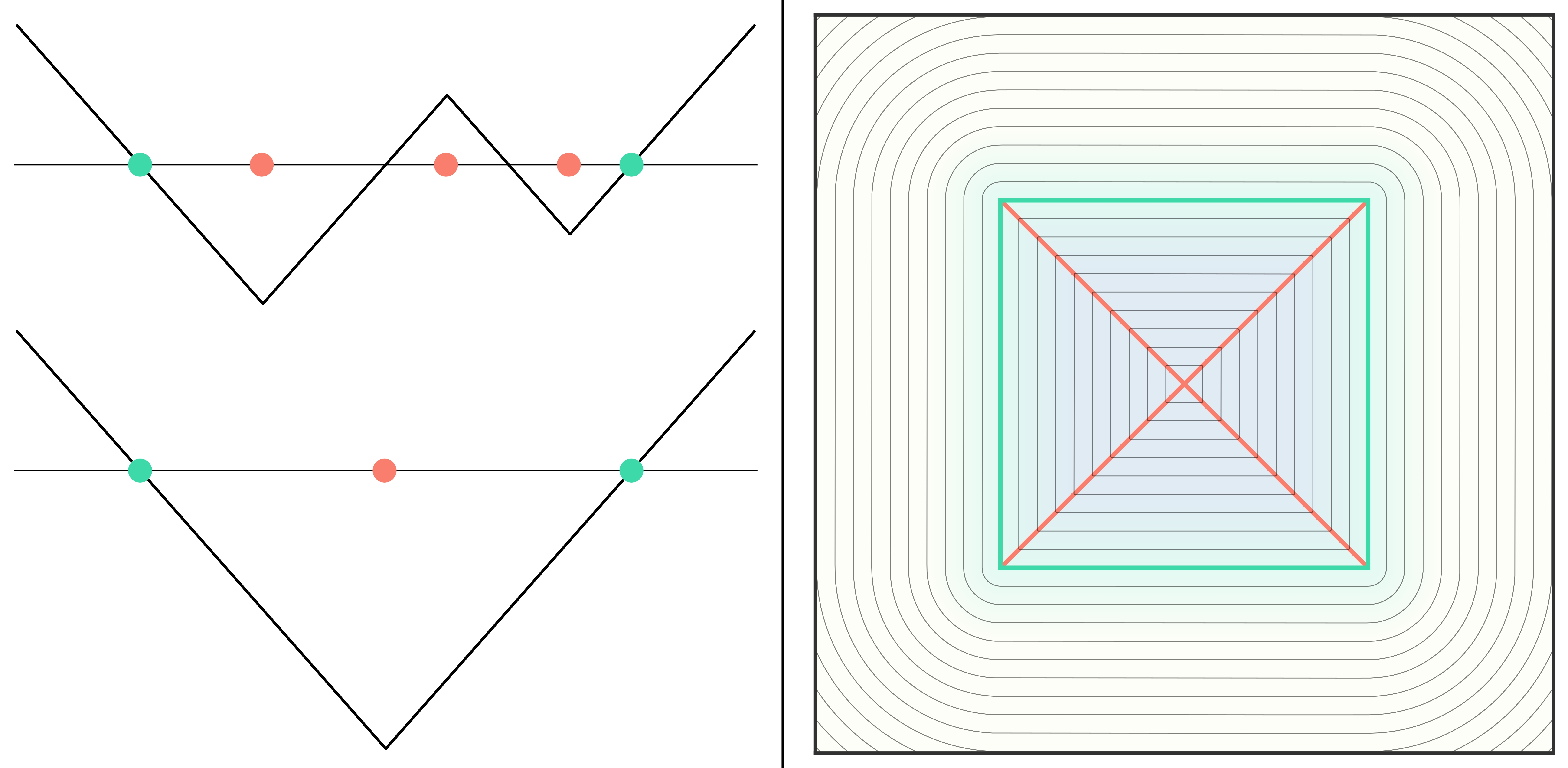}
\caption{Left: two graphs of one-dimensional functions satisfying the eikonal equation a.e., with zero boundary conditions at the green dots. The measure of the jump set (red dots) of the gradient $\jumpset$ is 3 at the top, and 1 for the viscosity solution displayed at the bottom. Right: SDF of a square (zero-level set in green) with the jump set $\jumpset$ (in red), also denoted as the medial axis. Note that the equally spaced levelsets indicate the constant slope of the SDF in normal direction.}
\label{fig:viscosity}
\end{figure}

At first,  we discuss different properties of the signed distance function of a closed, compact surface which 
will motivate the different components of our variational method.
We assume that the surface  $\surface = \partial\omega$ is the boundary of an open Lipschitz set $\omega\subset\Omega$ 
in a bounded computational domain $\Omega\subset \R^d$, with $d=2$ or $d=3$. Then, the signed distance function $\mathrm{sgndist}(\cdot,\surface)\colon \Omega\rightarrow\R$ of $\surface$ is defined as 
\begin{align*}
\mathrm{sgndist}(x,\surface)\coloneqq \begin{cases}
\mathrm{dist}(x,\surface)\quad&\text{, if }x\notin \omega \\
-\mathrm{dist}(x,\surface)\quad&\text{, if }x\in \omega\,. 
\end{cases}
\end{align*}
The SDF of $\surface$ is known to solve the eikonal equation as a function $\phi:\Omega\to\mathbb{R}$ with $\phi=0$ on $\surface$, and
\begin{align}
\label{eq:EikonalEq}
\|\nabla \phi\| = 1 \quad \text{a.e. in } \Omega.
\end{align}
The set of these solutions is very large in general.  
The concept of \emph{viscosity solutions} allows to identify the signed distance function as the unique weak solution $\phi$ of 
\cref{eq:EikonalEq} with $\phi=0$ on $\surface$ via a comparison principle, see \cref{fig:viscosity} for a 1D and a 2D example. This notion implies that the SDF is maximal in absolute values, compared to other solutions of \cref{eq:EikonalEq}.
To see this, let $x\in \Omega$, and $y\in \surface$, such that $\mathrm{dist}(x,\surface) = \|x-y\|$. Let $\phi:\Omega\to\mathbb{R}$ with $\phi=0$ on $\surface$ solve \cref{eq:EikonalEq}. Then we can estimate
\begin{align*}
	|\phi(x)| &= |\phi(x) - \phi(y)|\\ 
	&\leq \|\nabla \phi\|_\infty \|x-y\| = \|x-y\| = \mathrm{dist}(x,\surface)\,.
\end{align*}
Functions solving the eikonal equation possess an additional structural property.
In the direction normal to its level sets $\phi$ increases linearly, i.e.
\begin{align}
\label{eq:HessCondition}
D^2\phi\,\nabla \phi = 0
\end{align}
in regions where $\phi$ is twice differentiable.
This directly follows from a differentiation of $\|\nabla \phi\|^2 = 1$ and using the symmetry of the Hessian $D^2\phi$.
Another, geometric interpretation of \cref{eq:HessCondition} is that the gradient vector field $\nabla \phi$ 
of the signed distance $\phi$ is constant on rays pointing in the gradient direction, i.e.
\[0=\partial_t \left(\nabla\phi(x + t\nabla\phi(x))\right)\vert_{t=0} =D^2\phi(x)\nabla\phi(x)\,.\]
A key insight is that, while functions $\phi$ solving \cref{eq:EikonalEq} are Lipschitz and differentiable almost everywhere, 
their gradient  jumps on a set $\jumpset$, cf.~\cref{fig:viscosity}.
In fact, even for the SDF of a smooth surface $\surface$, the jump set of the gradient is a lower dimensional set of those points
where the nearest-point projection onto $\surface$ is not defined, also denoted as the medial axis of $\mathcal{S}$.
In \cite{wang2023neural} and \cite{hessian2}, the degeneracy condition of the Hessian \cref{eq:HessCondition} was used in the near-field of the surface to learn signed distance functions. Different from these approaches, our method takes advantage of  \cref{eq:HessCondition} globally and at the same time controls the co-dimension $1$ area of the jump set $\jumpset$.
\vspace{-0.2cm}
\section{Method}\label{sec:method}
We introduce a variational approach to compute the global SDF of a surface and its medial axis, i.e., the discontinuity set of the SDF's gradient.
The cost functional measures the  second order derivative in normal direction 
where the SDF is smooth. Furthermore, it controls the area measure of the discontinuity set.
\vspace{-0.2cm}
\subsection{Higher-order variational problem with jump term.}
In what follows, we describe the SDF as the minimizer of a variational problem that picks up the observations made in 
\cref{sec:motivation} in terms of cost functional components which 
\begin{enumerate}
\item[(i)] penalize deviations from the eikonal equation \cref{eq:EikonalEq},
\item[(ii)] regularize with a higher order term by enforcing the characteristic 
linearity along gradient directions away from the jump set $\jumpset$, cf. \cref{eq:HessCondition}, and 
\item[(iii)] control the measure of $\jumpset$ where $\nabla \phi$ is discontinuous.
\end{enumerate}

\noindent To this end, we take into account 
\[\int_\Omega \left\vert\|\nabla \phi\|^2 - 1\right\vert^2 \d x\]
as a penalty functional for the eikonal equation \cref{eq:EikonalEq}
and measure as a regularizer 
the second order derivative in normal direction in the square $L^2$ norm 
\[\int_{\Omega\setminus \jumpset}
\|D^2\phi\,\nabla\phi\|^2 \,\mathrm{d}x\,,\]
on the complement of $\jumpset$. 
Furthermore, we include  
the $(d-1)$-dimensional Hausdorff measure
\[\mathcal{H}^{d-1}(\jumpset)\,,\]
of the gradient jump set, which coincides for sufficiently regular $\phi$ with the length ($d=2$), or the area ($d=3$) of the medial axis, cf.~\cite{Fe69} for the definition.

The functional defined as a weighted sum of these three cost components is a higher-order analogue of the Mumford-Shah energy for image segmentation, cf. \cite{MuSh89}. The term $\mathcal{H}^{d-1}(\jumpset)$ corresponds to the discontinuity penalty 
for the jump of the image intensity, where we here evaluate the jump of the \emph{gradient} of $\phi$.
The term penalizing the second order derivative in normal direction replaces the Dirichlet energy of the image intensity. 
In both cases integration is performed only on the complement of the jump set. 
Finally, the eikonal loss plays the role of the fidelity term measuring the discrepancy of the denoised image from the noisy input. 

Let us emphasize that on the one hand the eikonal loss is a non-convex functional of $\nabla \phi$, where on the other hand
the highest order term is convex in the Hessian $D^2\phi$.

\begin{figure}[htbp!]
\includegraphics[width=\linewidth]{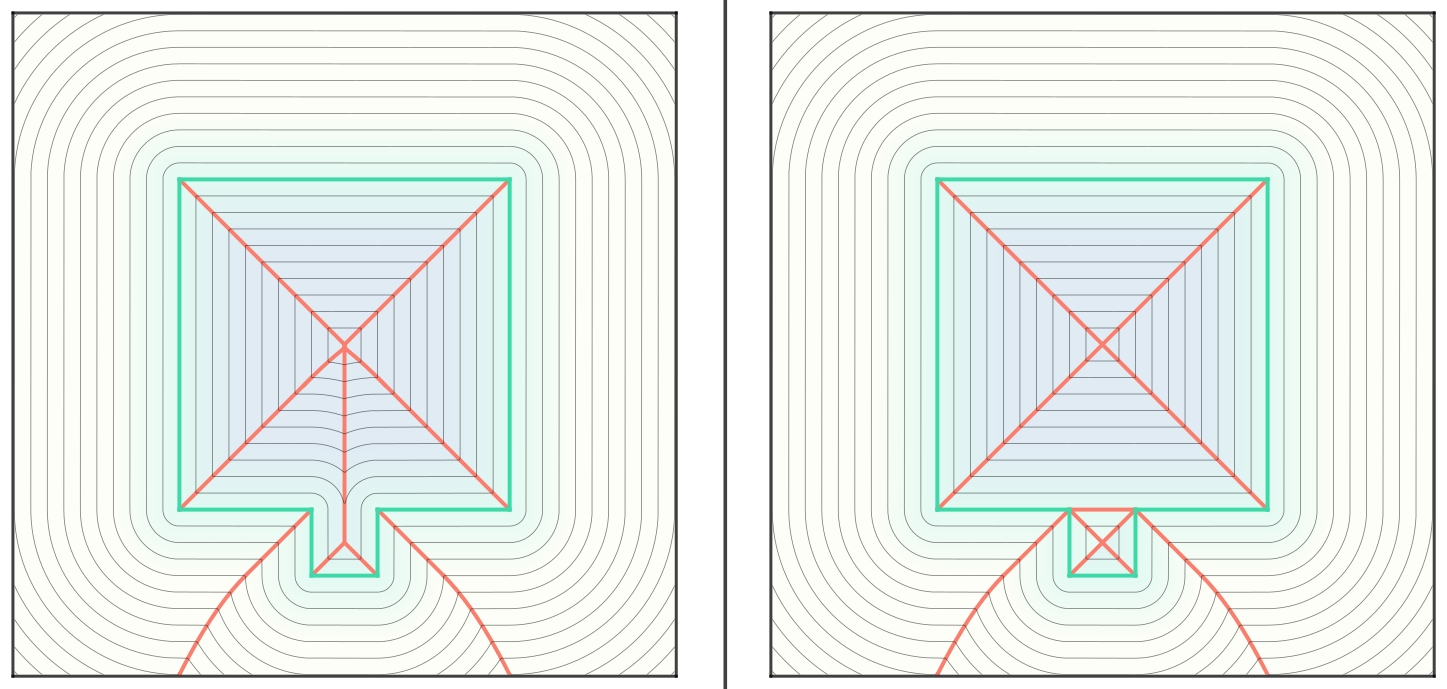}
\caption{Isolines of different solutions to the eikonal equation vanishing on the surface contour (in green) 
together with the gradient jump set (in red). Left: viscosity solution,
 Right: solution with shorter length of $\jumpset$ (example from \cite{AvGi96}).}
\label{fig:Aviles-GigaCounterexample}
\end{figure}

The functional with the above three components 
does not necessarily recover the \emph{viscosity} solution. In fact, there exist a.e. solutions of the Eikonal equation with smaller
jump set measure than the viscosity solution as depicted in \cref{fig:Aviles-GigaCounterexample}.

To increase the preference for the maximality property of the viscosity solution we add a further penalty 
$$
\mathcal{L}_\text{exp}[\phi] \coloneqq \sum_{p=1}^3\int_\Omega \exp(-\alpha_p |\phi|^p)\,\mathrm{d}x\,,
$$
favouring high values of $|\phi|$ for $\alpha_p>0$. A similar loss function was also used in \cite{SIREN} and \cite{wang2023neural}, where only $p=1$ was considered. The combination of three terms with $p=1,\,2,\,3$ and $\alpha_1=100$, $\alpha_2=100$, and $\alpha_3=10$ experimentally performed slightly better.
With the right scaling of $\mathcal{L}_\text{exp}$ and $\mathcal{H}^{d-1}(\jumpset)$, a minimizer of the resulting functional vanishing on $\surface$ is expected to be a signed distance function up to a factor $\pm 1$. 
Note that this scaling may depend on the surface $\surface$.

The identification of inside and outside is not required. In fact, the gradient of an unsigned distance function jumps on $\surface$, which leads to an avoidable increase of $\mathcal{H}^{d-1}(\jumpset)$.
\vspace{-0.3cm}
\subsection{Phase field approximation}
A direct discretization of the medial axis, with its generally complex topology is a serious challenge.
To handle such singularity sets phase field models provide a powerful alternative. 
Here, we pick up the phase field model introduced by Ambrosio and Tortorelli, cf. \cite{AmTo92}, 
for the approximation of the Mumford-Shah energy. 
Our model differs from the two-phase Modica-Mortola phase field model used in  \cite{Li21Phase}, 
which uses network based implicit surface representation and computes of a local approximation of the signed distance function in a postprocessing step.
In our approach, we apply the Ambrosio-Tortorelli model to represent the medial axis of the surface $\surface$ as an in general open discontinuity set of the SDF gradient.

At first, for a small parameter $\eps >0$, we introduce the scaled eikonal loss
\begin{align*}
\mathcal{L}_\text{eik}^\eps[\phi] \coloneqq\frac1\eps \int_\Omega \left\vert\|\nabla \phi\|^2 - 1\right\vert^2 \d x\,,
\end{align*}
which will imply the convergence of  $\phi$ with bounded loss
to a solution of \cref{eq:EikonalEq} for $\eps$ tending to $0$.
Then, we introduce the higher order regularization functional
\begin{align*}
\mathcal{L}_\text{HO}^\eps[\phi,v]\coloneqq \int_\Omega v^2\|D^2\phi\nabla \phi\|^2 + \eps^2\|D^2 \phi\|^2 \d x
\end{align*}
together with the Ambrosio-Tortorelli loss functional
\begin{align*}
\mathcal{L}_\text{AT}^\eps[v] \coloneqq \int_\Omega \eps \|\nabla v\|^2 + \frac{1}{4\eps}(v-1)^2 \d x\,,
\end{align*}
where $v: \Omega \to \R$ is a phase field function with $v \approx 1$ away from the jump set and
decaying significantly in a transition region of width $\eps$ around the jump set.
For fixed $\eps$ the first term in  $\mathcal{L}_\text{AT}^\eps$ fosters smoothness of the phase field, whereas the second term 
expresses a preference for $v\approx 1$ on sets of positive Lebesgue measure. 
For $\eps\to 0$ and minimizing tuples $(\phi^\eps,v^\eps)$  
the associated Ambrosio-Tortorelli loss $\mathcal{L}_\text{AT}^\eps[v^\eps]$  
is expected to converge to the $\mathcal{H}^{d-1}$-area of the gradient jump set of the limit of  $\phi^\eps$, due to the coupling induced by $\mathcal{L}_\text{HO}^\eps[\phi,v]$.
Small values of $v^\eps$ allow for local regularity defects in $D^2\phi^\eps\nabla \phi^\eps$ 
close to the jump set. A weighting of the eikonal loss with $\frac1\eps$ fosters a thin representation of the medial axis, with a preferred width decaying faster than linear in $\eps$. Compared to the higher-order term, a scaling with $v^\eps$ is not required, because the intergrand is uniformly bounded. 
In fact, the higher order regularization is active only in direction of the gradient of the SDF approximation and only away from the jump set. 
To render the phase field model for fixed $\eps>0$ wellposed, we add the second term in $\mathcal{L}_\text{HO}^\eps$ as an 
isotropic regularizer, uniformly weighted with $\eps^2$.

Finally, we have to ensure for the SDF that $\phi\approx 0$ on $\surface$. 
This is resembled by the reconstruction loss
\begin{align*}
\mathcal{L}_\text{recon}^\eps[\phi] \coloneqq \frac{1}{\eps^2} \int_\surface \phi^2 \d \mathcal{H}^{d-1}\,.
\end{align*}

Altogether, we obtain the total loss functional  
\begin{align} 
\mathcal{L}_\text{total}^\eps[\phi,v] \coloneqq
&  \gamma_\text{HO}\mathcal{L}_\text{HO}^\eps[\phi,v] 
+ \gamma_\text{AT}\mathcal{L}_\text{AT}^\eps[v] 
+ \gamma_\text{recon}\mathcal{L}_\text{recon}^\eps[\phi] \nonumber\\ 
&+ \gamma_\text{eik}\mathcal{L}_\text{eik}^\eps[\phi] 
+ \gamma_\text{exp}\mathcal{L}_\text{exp}[\phi] \,, \label{eq:totalLoss}
\end{align}
with positive weights $\gamma_\text{HO}, \gamma_\text{AT}, \gamma_\mathrm{recon}, \gamma_\mathrm{eik}, \gamma_\text{exp}>0$.

The following theorem establishes the existence of minimizing tuples $(\phi^\eps,v^\eps)$ for $\eps>0$.
\begin{theorem}[Existence of minimizers of $\mathcal{L}_\text{total}^\eps$]

\label{thm:ExistenceOfMinimizers}
Let the surface $\surface$ be the boundary of an open Lipschitz set $\omega\subset\Omega$ 
in the bounded computational domain $\Omega \subset \R^d$, with $d=2$ or $d=3$. Then, for 
fixed $\eps >0$, there exists a minimizing tuple
$(\phi^\eps,v^\eps)\in H^2(\Omega)\times H^{d-1}(\Omega)$ of $\mathcal{L}_\text{total}^\eps[\cdot,\cdot]\,$.
\end{theorem}
Here, $H^s(\Omega)$ is the space of functions with square integrable weak derivatives of order up to $s$.
The proof can be found in \cref{sec:appproofs}.

Next, we consider the consistency of the phase field energy with the limit energy for $\eps \to 0$. 
To this end, let $(\jumpset)_{\delta} = \{x\in \Omega\,:\; \mathrm{dist}(x,\jumpset) < \delta\}$ 
be the tubular neighborhood of $\jumpset$ with radius $\delta$, and let 
$|(\jumpset)_{\delta}|$ be it's $d$-dimensional Lebesgue measure.

\begin{theorem}{(Recovery sequence)}
	\label{thm:limsup}
	Let $\phi \in W^{1,2}(\Omega)$, with $\phi = 0$ on $\surface$ and $\|\nabla \phi\|= 1$ almost everywhere. Assume additionally that $\lim\limits_{\delta\rightarrow 0^+}\frac{|(\jumpset)_{\delta}|}{2\delta} = \mathcal{H}^1(\jumpset)<\infty$. Then there exists a sequence $(\phi_\eps,v_\eps)_\eps \subset W^{2,2}(\Omega)\times W^{1,2}(\Omega;[0,1])$ such that $\phi_\eps \rightarrow \phi$ in $W^{1,2}(\Omega)$, and $v_\eps\rightarrow 1$ in $L^2(\Omega)$, and 
	\begin{align*}
	\lim_{\eps\rightarrow 0} \mathcal{L}_\text{HO}^\eps[\phi^\eps,v^\eps] &= 
	\int_{\Omega\setminus \jumpset} \left\Vert D^2\phi\,\nabla\phi\right\Vert^2 \,\d x = 0\,, \\
	\lim_{\eps\rightarrow 0} \mathcal{L}_\text{exp}[\phi^\eps] &= \mathcal{L}_\text{exp}[\phi]\,,\\
	\lim_{\eps\rightarrow 0} \mathcal{L}_\text{AT}^\eps[v^\eps] &= \mathcal{H}^{d-1}(\jumpset)\,,\\
	\lim_{\eps\rightarrow 0} \mathcal{L}_\text{eik}^\eps[\phi^\eps] &= 0\,,\quad
	\lim_{\eps\rightarrow 0} \mathcal{L}_\text{recon}^\eps[\phi^\eps] = 0\,.
	\end{align*}
\end{theorem}
The proof can be found in \cref{sec:RecoverySequence}.

For the phase field approximation of the Mumford-Shah model the construction of a recovery sequence can be complemented to a full 
$\Gamma$-convergence result (cf.~\cite{AmTo92}). For our model, the associated $\liminf$ inequality relating the phase field functional 
$\mathcal{L}_\text{HO}^\eps$ and the limit functional 
$\int_{\Omega\setminus \jumpset} \left\Vert D^2\phi\,\nabla\phi\right\Vert^2 \,\d x$ as well as the recovery sequence for general 
eikonal solutions is still open. 

\section{Implementation}
Throughout all experiments, we consider surfaces $\surface$, represented by unoriented point clouds contained in the computational domain $\Omega\coloneqq[-1.2,1.2]^d$. For the phase field parameter $\eps$ we choose $\eps=10^{-3}$ for $d=2$ and $\eps=10^{-4}$ for $d=3$.
We implement our method using the Pytorch framework, cf. \cite{PyTorch}. All experiments were performed on a workstation with two AMD
EPYC 7402 processors, 256GB RAM, and an NVIDIA A100 with
40GB memory. 
The code is available at \url{https://github.com/sweidemaier/Neat_SDF}.
\vspace{-0.3cm}
\subsection{Networks}
To represent the SDF $\phi$ and the phase field $v$, we use two separate networks $\phi_\theta$ and $v_\eta$ with parameter vectors $\theta$ and $\eta$. The input to both networks is the spatial coordinate $x\in \Omega$. 

For the phase field network $v_\eta$, we employ a ResNet architecture, cf. \cite{ResNet}, with sinusoidal activation functions, referred to as SIREN activations (cf. \cite{SIREN}). This combination allows for a multi-scale type of approximation, especially suited for the highly localized phase field profile.
To enforce the phase field to remain in $[0,1]$ without overshooting, we augment the final layer by concatenating the ResNet output with a sigmoid function $\tfrac{e^{\delta x}}{1 + e^{\delta x}}$
with $\delta=0.1$, representing a smooth transitions from $0$ as $x \to -\infty$ to $1$ as $x \to \infty$. For 2D experiments, we use 4 ResNet blocks with 64 hidden units each, and for 3D, we use 6 ResNet blocks with 128 hidden units each.

For the SDF network we employ a quadratic neural network (QuaNet, cf. \cite{QuaNets}). Concretely, the SDF $\phi_\theta$ is parameterized by an MLP with 256 hidden units per layer, where we use 4 layers for 2D and 8 layers for the 3D experiments. Each hidden layer implements a quadratic expansion of its pre-activation followed by the smooth SIREN activation. The combination of the QuaNet architecture with SIREN activation was also used by \cite{HotSpot} for the computation of SDFs.

We optimize using stochastic gradient descent simultaneously for the phase field and SDF network parameters using the Adam approach \cite{Kingma2014AdamAM} with different gradient descent parameters.
Specifically, we use $\beta_1 = 0.9$ and $\beta_2 = 0.999$ for the phase field network, and $\beta_1 = 0.9$ and $\beta_2 = 0.98$ for the SDF network. The learning rates are set to $5\times 10^{-4}$ for the phase field network and $5\times 10^{-5}$ for the SDF network, and are decreased during training.

\vspace{-0.4cm}
\subsection{Adaptive Sampling}
\begin{figure}
\includegraphics[width=\linewidth]{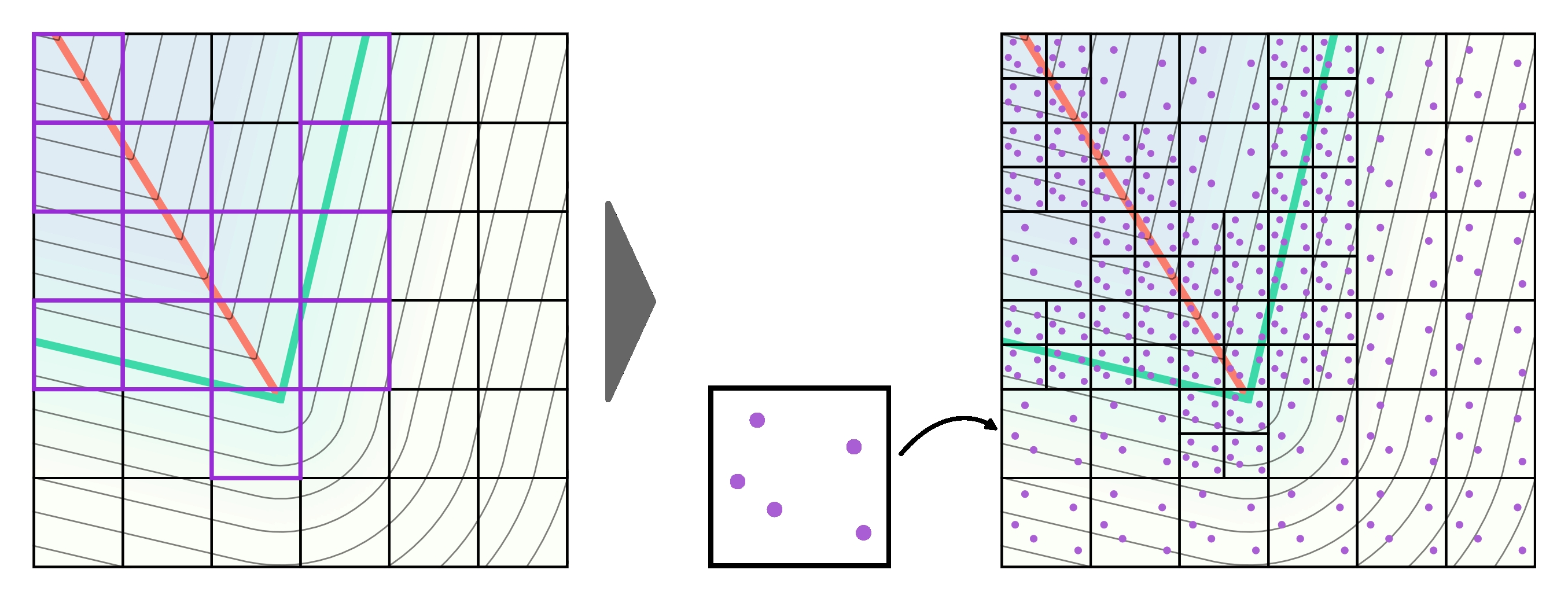}
\caption{Left: regular grid, where cells with $|\phi_\theta| 
< \tau_{\mathrm{sdf}}$ or $v_\eta < \tau_{\mathrm{pf}}$ are marked. Right: Subdivision of marked cells and sampling of quadrature points on reference cell.}
	\label{fig:adaptive_sampling}
\end{figure}

To approximate the volume integrals in the loss functional, we employ a Monte Carlo quadrature scheme and use an adaptive sampling strategy during training of 
our 3D models. In particular, we sample more points close to the approximate medial axis and the surface with suitably adapted integration weights, to achieve 
a high resolution of the phase field close to the jump set $\jumpsetTheta$ and a good reconstruction of the surface $\surface$. This overall results in an improved approximation of the global signed distance function.
For this we utilize a simple and yet effective sampling strategy, based on local grid refinement, with refinement level $i\in\N$, in regions with $|\phi_\theta| < 2^{-i}\tau_{\mathrm{sdf}}$ or $v_\eta < \tau_{\mathrm{pf}}$, where $\tau_{\mathrm{sdf}}$ and $\tau_{\mathrm{pf}}$ are given thresholds. 
In practice we use $\tau_{\mathrm{sdf}} = 0.1$ and $\tau_{\mathrm{pf}} = 0.75$. 
These regions are identified in a preprocessing step by dense uniform sampling.
A sketch, displaying this adaptive sampling strategy, is shown in \cref{fig:adaptive_sampling}. The detailed algorithm for one training based on adaptive sampling is explained in \cref{alg:adaptive_sampling}. 
The weights associated with the sample points in cells on refinement level $i$ are $(2^{i\,d}n)^{-1}h^d$. Here, $h>0$ is the initial grid size and $n\in\N$ is the number of sample points in the cell.
The training of the parameter vectors $\theta$ and $\eta$ for the networks $\phi_\theta$ and $v_\theta$
is repeated until a stopping criterion is met. For our experiments, we trained for a total of 30 epochs.

Furthermore, to accurately approximate the reconstruction loss $\mathcal{L}_\text{recon}^\eps$, we employ the local weighting scheme introduced in \cite[section 3.1.1]{HeatSDF26}, where the weights are increased where the point cloud is sparse.
 
\begin{algorithm}[t]
\caption{Training based on adaptive sampling}
\label{alg:adaptive_sampling}
\begin{algorithmic}[1]
\Require Batch sizes \(N\), \(M\), s.t. \(N \ll M\), initial grid size \({h\coloneqq|\Omega|^{\frac1d}m^{-1}}\), thresholds \(\tau_{\mathrm{sdf}}, \tau_{\mathrm{pf}}\), grid depth k.  

\State /* Grid generation */

\State Initialize uniform rectangular grid \(\mathcal{G}_0\) on $\Omega$ with cell size \(h\)

\State Sample test set \(\mathcal{P} \subset \Omega\) with \(|\mathcal{P}| = M\)
\For{$i = 0$ to $k-1$}
\State \(\mathcal{G}_{i+1} = \mathcal{G}_i\)
\For {all cells \(C \in \mathcal{G}_{i}\)}
    \For {\(x\in \mathcal{P}\cap C\)}
        \If{\( |\phi_\theta(x)| < {2^{-i}} {\tau_{\mathrm{sdf}}} \) \textbf{or} \(v_\eta(x) < {\tau_{\mathrm{pf}}}\)}
            \State Update $\mathcal{G}_{i+1}$ by a uniform subdivision of $C$
            \State into $2^d$ subcells
        \EndIf
    \EndFor
\EndFor
\EndFor     

\State /* Training */

\State Let \(M_k = |\mathcal{G}_k|\) be the number of grid cells
    \State \(n \gets \left\lceil \frac{N}{M_k} \right\rceil\)
    \State Initialize empty batch \(\mathcal{B}\)
    \For{each cell \(C \in \mathcal{G}_k\)}
        \State Sample \(n\) random points in \([0,1]^d\)
        \State Scale and shift points to lie inside \(C\)
        \State Add points to batch \(\mathcal{B}\)
    \EndFor
    \State Perform training using batch \(\mathcal{B}\)
\end{algorithmic}
\end{algorithm}

\subsection{Loss term scheduling}
\label{sec:Scheduling}
For both 2D and 3D experiments, we employ a three-phase training schedule, where the weights in \cref{eq:totalLoss} are adapted during training. The starting and final weights are listed in \cref{Appendix:2D} and \cref{Appendix:3D}.

\paragraph*{Phase 1 (SDF Initialization).}
We start training to obtain a proper first approximation of the  SDF and some approximate 
reconstruction of the surface as its zero-level set. As an initialization, we choose an approximate SDF of the unit sphere, i.e. $\phi_\theta(x) \approx \Vert x\Vert - 1$. 
During this phase, we set $\gamma_{\text{exp}}$ to a high value to encourage SDF growth away from the surface, while keeping $\gamma_{\text{eik}}$ and $\gamma_{\text{HO}}$ relatively low and thereby eliminating ghost geometry. The phase field $v\equiv 1$ is held fixed in this phase, and we solely train the SDF. 

\paragraph*{Phase 2 (Joint SDF and Phase Field Optimization).}
After a fixed number of training epochs (in our implementation $5$), we activate the phase field optimization and stepwise increase $\gamma_{\text{eik}}$ and $\gamma_{\text{HO}}$ up to the destinated values.
Simultaneously, we decrease $\gamma_{\text{exp}}$ down to the destinated value.
The phase field is generally easier to train and we observe a relatively quick convergence.

\paragraph*{Phase 3 (SDF Refinement).}
In the final phase (in our implementation after 20 epochs) 
we freeze the phase field network and continue optimizing the SDF. 
This allows for a fine-tuning of the distance function in light of the fixed phase field representing the discontinuity set while avoiding unnecessary computational overhead from further phase field updates. In this phase we additionally evaluate the second-order term $v_\eta^2 \|D^2\phi_\theta\nabla \phi_\theta\|^2 $ on the point cloud.

This scheduling strategy has shown to ensure stable convergence in application. It
leverages the complementary roles of the eikonal constraint, higher-order regularization, 
and phase field approximation throughout the training process.
\vspace{-0.3cm}
\section{Experiments}
\subsection{Experiments in 2D}
We test our method on the two-dimensional dataset provided by \cite{HotSpot}. The computed SDFs and phase fields corresponding to this dataset are shown in \cref{fig:2dExperiments} in \cref{Appendix:2D}, together with the ground truth solution for a subset, showing good agreement of our SDF solution, and of the medial axis approximated by the phase field.

\begin{figure}[htbp!]
	\centering
    {\includegraphics[width=0.95\linewidth]{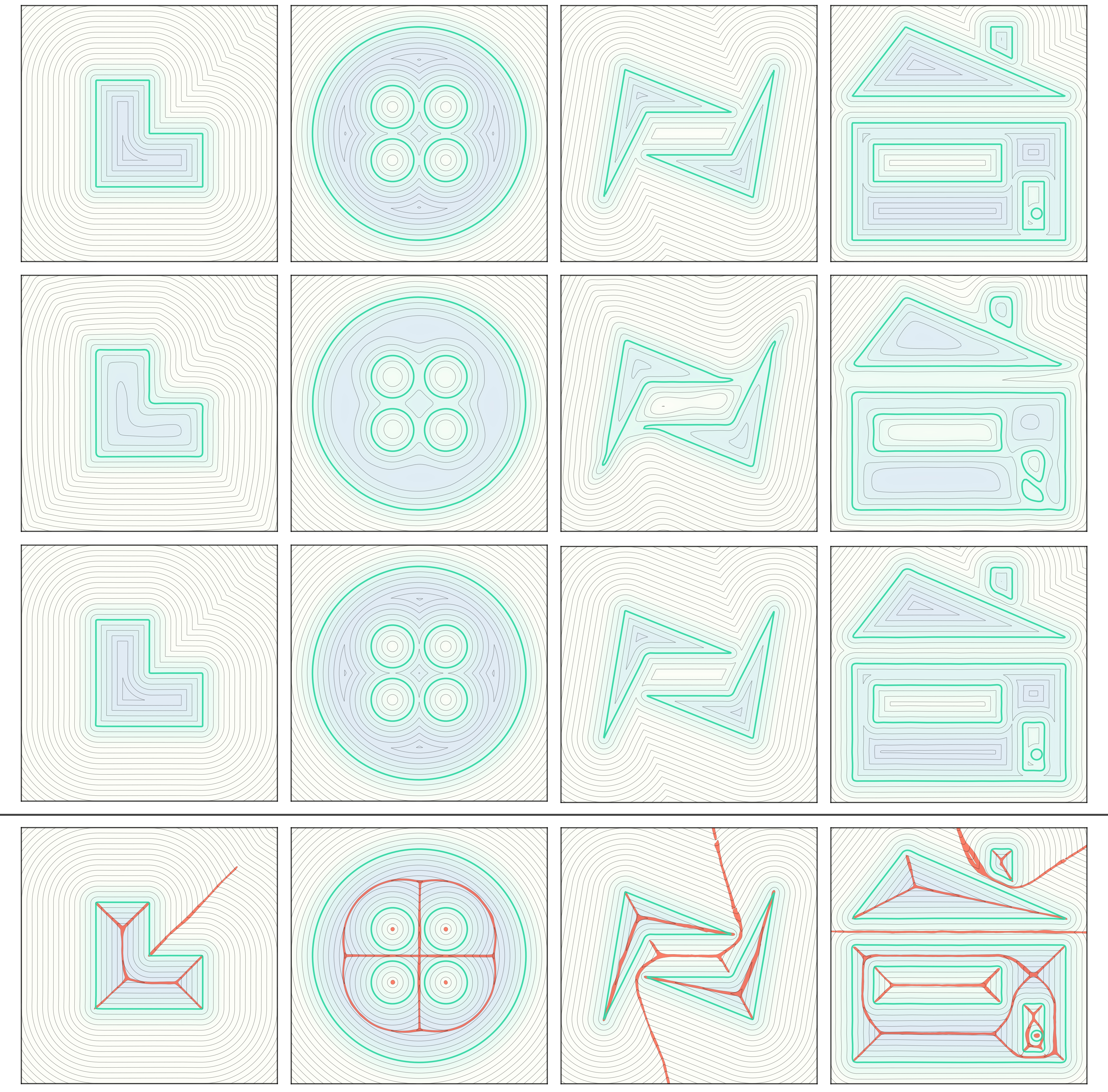}}
    \begin{tikzpicture}
        \centering
    \tiny{
    \rotatebox{90}{
        \node[font=\bfseries\small] at (0, 1.6) { };
        \node[font=\bfseries\tiny] at (2.7, -0.1) {Ours (SDF \& PF)};
        \node[font=\bfseries\tiny] at (4.7, -0.1) {Ours (only SDF)};
        \node[font=\bfseries\tiny] at (6.7, -0.1) {HotSpot};
        \node[font=\bfseries\tiny] at (8.6, -0.1) {Ground truth};
    }}
\end{tikzpicture}
\vspace{-2.4cm}\caption{From top to bottom: Ground truth SDF solution, Computed SDF using the Hotspot method \cite{HotSpot}, Computed SDF using our method, Computed SDF using our method with the $0.25$-sublevel set of the phase field (PF, in red).}

\label{fig:2dComparisonHotspot}
\end{figure}

In \cref{fig:2dComparisonHotspot}, we display a visual comparison of the computed SDF of four of the 2D-shapes of the Hotspot method \cite{HotSpot} and our method, compared to the ground truth. Especially inside the shapes, our method shows better distance computation visually, compared to Hotspot. The deviation from the ground truth is localized at very few spots such as the door of the house contour on the right.
\vspace{-0.3cm}
\subsection{Experiments in 3D}

\subsubsection{Evaluation metrics }
Both high-detail surface reconstruction and accurate distance estimation are crucial for a good SDF approximation. We employ different metrics to evaluate the performance of our method with respect to these two aspects. 
\\First, to measure the surface reconstruction error, we compute the Chamfer distance 
\begin{equation*}
\mathbf{d_{C}} \coloneqq \frac{1}{|\mathcal{P}|}\sum_{x\in \mathcal{P}} \min_{y\in \hat{\mathcal{P}}} \|x-y\| + \frac{1}{|\hat{\mathcal{P}}|}\sum_{y\in \hat{\mathcal{P}}} \min_{x\in \mathcal{P}} \|x-y\|,
\end{equation*}

Additionally, we compute the Hausdorff distance
\begin{equation*}
\mathbf{d_{H}} \coloneqq \max\left\{\max_{x\in \mathcal{P}} \min_{y\in \hat{\mathcal{P}}} \|x-y\|, \max_{y\in \hat{\mathcal{P}}} \min_{x\in \mathcal{P}} \|x-y\|\right\}\,.
\end{equation*}
For both Hausdorff and Chamfer distance, we use two point clouds $\mathcal{P}$ and $\hat{\mathcal{P}}$ with $100$k points each, sampled from the ground truth surface $\surface$ and the zero-level set $\hat \surface$ of the computed SDF, respectively. To extract the zero-level set of the computed SDF, we use Marching Cubes with high resolution (in our implementation $512^3$).
Further, we evaluate the normal alignment error on the input surface by computing 
\begin{equation*}
\mathbf{E_n} \coloneqq 1 - \frac{1}{|\mathcal{F}|} \sum_{F\in \mathcal{F}} n(x_F)\cdot\frac{\nabla  \phi(x_F)}{\|\nabla \phi(x_F)\|}\,, 
\end{equation*}
where $n(x_F)$ denotes the discrete normal of the respective ground truth mesh at the triangle center $x_F$ 
of the triangle $F\in\mathcal{F}$, with  $\mathcal{F}$ denoting the set of all triangles, and $|\mathcal{F}|$ 
the number of triangles, respectively.
To this end, we exploit that all considered unoriented point clouds come with  ground truth meshes.

We further evaluate the quality of the results inside the whole computational domain $\Omega=[-1.2, 1.2]^3$ using a point cloud $\mathcal{P}_\Omega$ of $50$k uniformly sampled points and on a point cloud $\mathcal{P}_\mathcal{N}$ of $10$k points in a narrow band $\mathcal{N}$ with distances in $[-0.1,0.1]$ from the ground truth surface $\surface$. 
The set $\mathcal{N}$ was generated once per model using rejection sampling based on a mesh-based pointwise distance computation with high accuracy \cite{MeshToSDF} and was subsequently used consistently across all evaluations. This method also serves as a ground truth signed distance  
$\mathrm{sgndist}(\cdot, \mathcal{S})$ to the mesh surface $\mathcal{S}$, which we use for comparison to calculate the SDF error on the whole domain $\Omega$ and the narrow band $\mathcal{N}$, respectively:
 \begin{align*}
\mathbf{E}_\mathbf{SDF}^{\Omega} &\coloneqq 
\sqrt{
	\frac{1}{|\mathcal{P}_\Omega|}\sum_{x\in \mathcal{P}_\Omega}|\phi(x) - \mathrm{sgndist}(x, \mathcal{S})|^2
}\,,
\\
\mathbf{E}_\mathbf{SDF}^\mathcal{N} &\coloneqq \sqrt{\frac{1}{|\mathcal{P}_\mathcal{N}|}\sum_{x\in \mathcal{P}_\mathcal{N}}|\phi(x) - \mathrm{sgndist}(x, \mathcal{S})|^2}\,,
\end{align*}
as well as the eikonal error: 
 \begin{align*}
\mathbf{E}_\mathbf{eik}^\Omega &\coloneqq   \frac{1}{|\mathcal{P}_\Omega|}\sum_{x\in \mathcal{P}_\Omega} \left\vert 1 - \|\nabla\phi(x)\|\right\vert\,,\\
\mathbf{E}_\mathbf{eik}^\mathcal{N} &\coloneqq   \frac{1}{|\mathcal{P}_\mathcal{N}|}\sum_{x\in \mathcal{P}_\mathcal{N}} |1 - \|\nabla\phi(x)\|| \,.
\end{align*}

\subsubsection{Comparison}
For the three-dimensional case, we compare our method against 5 state-of-the-art methods for neural SDF computation, including the neural network based methods HeatSDF \cite{HeatSDF26}, Hessian \cite{wang2023neural}, HotSpot \cite{HotSpot}, and 1-Lip \cite{coiffier20241}. For the HeatSDF method, we used the version of the code that uses Winding numbers for inside/outside segmentation. \\
Furthermore, we compare to a grid-based method, the generalized signed distance method (GSD) \cite{GSD}. All experiments with GSD were carried out with a grid size of $128^3$ ($\approx$ 2 million nodes). For comparison, our method uses a combined number of degrees of freedom for SDF and phase field of about $85^3 $ ($\approx 600\mathrm{k}$).\\
For each of those methods we use the standard implementation provided by the respective authors.
Note that the methods 1-Lip, HeatSDF, and GSD require oriented point clouds or surface normals as input. For 1-Lip and HeatSDF, the normal information is only used in a preprocessing step to determine inside and outside regions using generalized winding numbers, while GSD diffuses the input normals to obtain a smooth normal field. Our method does not require any additional information to construct SDFs.
 Note that the 1-Lip method enforces the solution to have gradient norm smaller or equal to one.

\begin{figure*}[htbp!]
	\centering
	\begin{tikzpicture}
	\node[inner sep=0pt] at (0.0, 0.0) {\includegraphics[width=\linewidth]{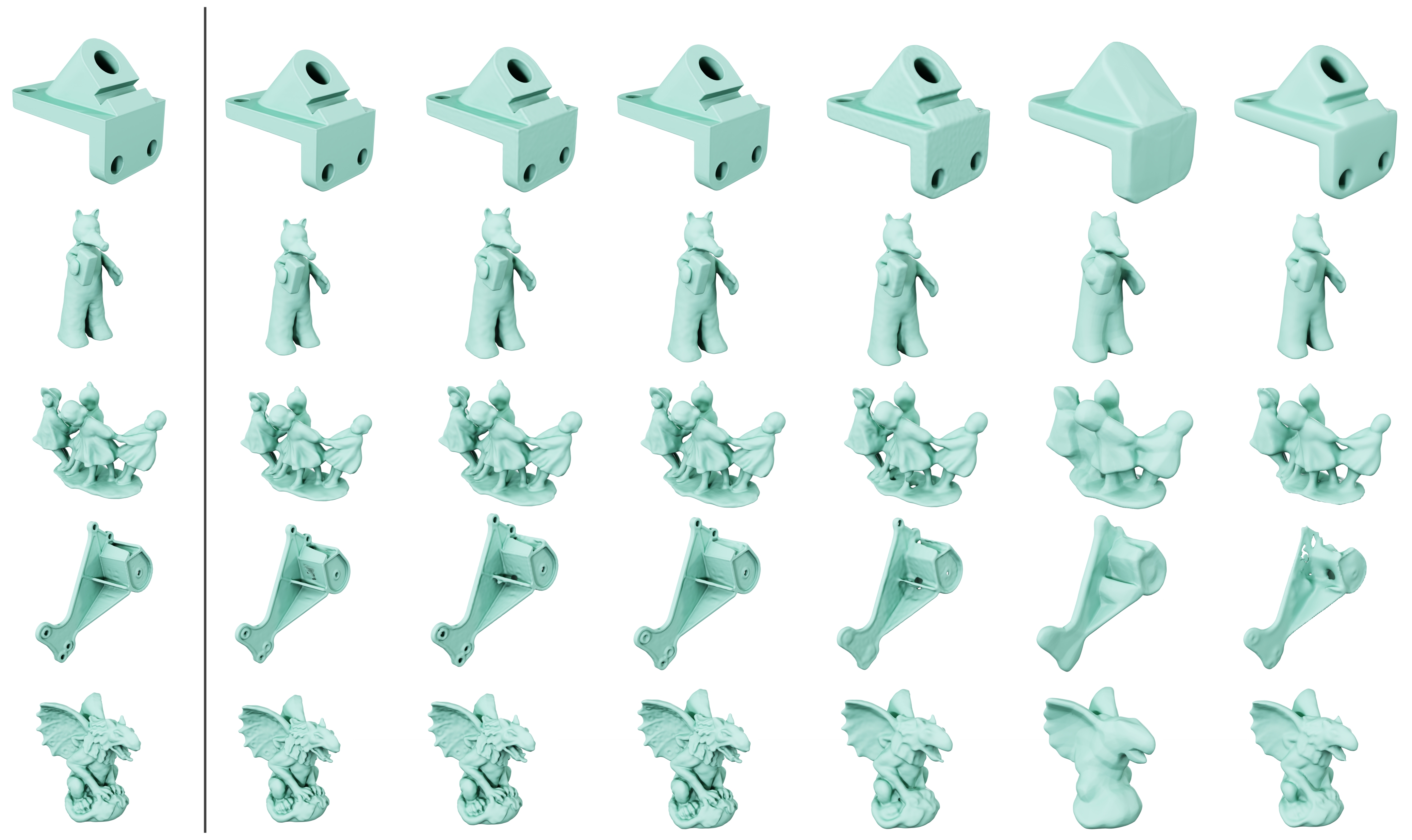}};
	\node[inner sep=0pt] at (-7.7, 5.1) {\textbf{Ground truth}};
	\node[inner sep=0pt] at (-5.1, 5.1) {\textbf{Ours}};
	\node[inner sep=0pt] at (-2.5, 5.1) {\textbf{HotSpot}};
	\node[inner sep=0pt] at (-0.1, 5.1) {\textbf{Hessian}};
	\node[inner sep=0pt] at (2.5, 5.1) {\textbf{HeatSDF}};
	\node[inner sep=0pt] at (5.1, 5.1) {\textbf{1-Lip}};
	\node[inner sep=0pt] at (7.7, 5.1) {\textbf{GSD}};
	\end{tikzpicture}
	
	\caption{Results for the SRB dataset for five different neural SDF reconstruction methods and one grid-based method (GSD). Across all five shapes, the zero-level sets of the Hessian method and Ours are closest to the groundtruth geometry (left), showing both high details and a good reconstruction of flat regions. The HotSpot method captures a similar level of details, while struggling with flat regions and sharp edges (see, e.g. top row example). The remaining methods (HeatSDF, 1-Lip, GSD), show a significant smoothing of the geometry and miss details.
	}
	\label{fig:SRB_vis}
\end{figure*}

We first evaluate all methods on the surface-reconstruction-benchmark dataset (SRB, cf. \cite{SRB_dataset}), see \cref{fig:SRB_vis}, \cref{tab:srb_results_surf} and \cref{tab:combined}. Our method achieves high accuracy in regions with fine detail, while also reconstructing flat regions with almost no perturbation, and properly resembling sharp edges with jumping normal.
Quantitatively, our approach ranks consistently among the top methods for both surface reconstruction (\cref{tab:srb_results_surf}) and SDF accuracy metrics (\cref{tab:combined}).

\begin{table}[htbp!]
	\centering
	
	\setlength{\tabcolsep}{2.5pt}
	\begin{tabular}{
			l
			cc
			cc
			cc
		}
		\toprule
		
		& \multicolumn{2}{c}{$\mathbf{d_{C}}$}
		& \multicolumn{2}{c}{$\mathbf{d_{H}}$}
		& \multicolumn{2}{c}{$\mathbf{E_{n}}$} \\
		\cmidrule(lr){2-7} 
		\textbf{Method}
		& mean & std.
		& mean & std.
		& mean & std. \\
		
		\midrule
		\textbf{GSD} $(\star)$& 0.0302 &0.0074 & 0.1171 & 0.0450  & 0.1249 & 0.1736 \\
		\textbf{1-Lip}        & 0.0573 & 0.0170   & 0.2468 & 0.0965 &  0.1188 & 0.0332  \\
		\textbf{HeatSDF}  & 0.0147 & 0.0030  & 0.1521 & 0.0609& 0.0378 & 0.0255 \\
		\textbf{Hessian}      & \textbf{0.0102} & \underline{0.0016}  & \textbf{0.0521} & \textbf{0.0327}  & {0.0116} & 0.0126  \\
		\textbf{HotSpot}       & 0.0105 & 0.0017 & 0.0704 & 0.0427  & \underline{0.0114} & \textbf{0.0079} \\
		\midrule
		\textbf{Ours}         & \underline{0.0104} & \textbf{0.0014 }& \underline{0.0645} & \underline{0.0329}  & \textbf{0.0110} & \underline{0.0088}   \\
		\bottomrule
	\end{tabular}
	\caption{Averaged surface reconstruction metrics for the SRB dataset. Mean and standard deviation are reported across all shapes. The best results are highlighted in bold and the second-best results are underlined. $(\star)$ denotes a grid based method.}
	\label{tab:srb_results_surf}
\end{table}

In terms of computational efficiency, our method is competitive with existing approaches, though it simultaneously optimizes both the SDF and the medial axis approximation via the phase field. Further details on runtime comparisons are provided in \cref{tab:combined}.

Additionally, we compare results for a total of 40 shapes from the Thingy10k dataset \cite{Thingi10K},
where 20 shapes are randomly sampled from the full dataset and 20 shapes are sampled from the pool of shapes with the tags 'sculpture' or 'scan'. 
The resulting selection of shapes is especially challenging for SDF reconstruction, including shapes with thin structures, high curvature, and flat regions.
In \cref{fig:thingy10k_vis} in \cref{Appendix:3D} qualitative results for a subset of these shapes is presented. Our method achieves high-quality surface reconstruction and accurate signed distance estimation across both local and global error metrics. Corresponding quantitative results are displayed in \cref{tab:thingy10k_results}. 

\begin{figure}[h]
	\centering
	\includegraphics[scale=0.4]{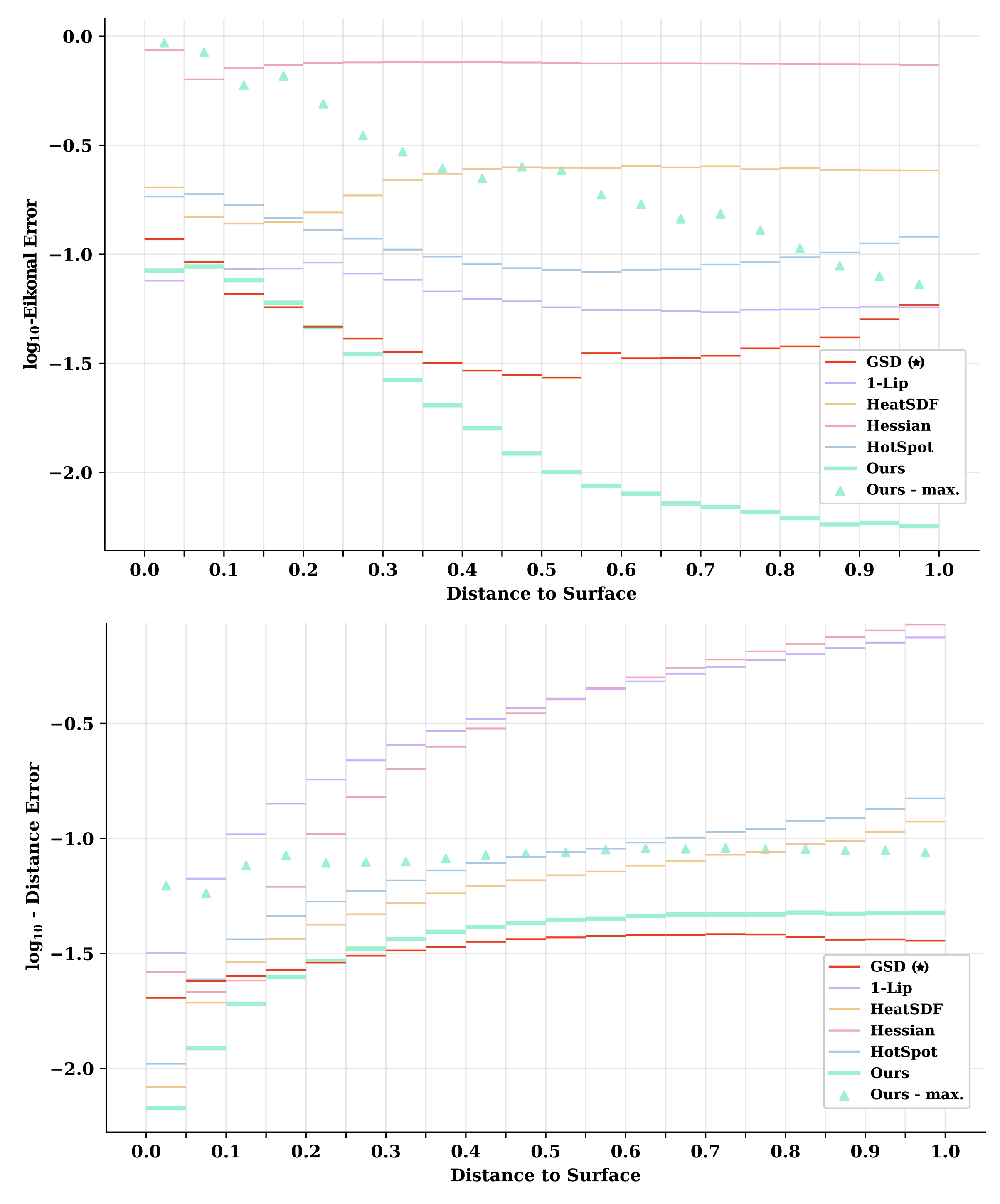}
	\caption{Quantitative evaluation of the Eikonal error (top) and the SDF error (bottom) averaged across the five shapes from the SRB dataset and evaluated on ground truth distance bands of width $0.05$ to the surface (horizontal axes). Triangles denote the maximum error on the respective narrow band. $(\star)$ denotes a grid based method.}
	\label{fig:quantComp}
\end{figure}

\begin{table*}[htbp!]
	\centering
	\begin{tabular}{
			l
			cc
			cc
			cc
			cc
			c
		}
		\toprule
		
		& \multicolumn{2}{c}{$\mathbf{E}_\mathbf{eik}^\Omega$}
		& \multicolumn{2}{c}{$\mathbf{E}_\mathbf{eik}^\mathcal{N}$}
		& \multicolumn{2}{c}{$\mathbf{E}_\mathbf{SDF}^{\Omega}$}
		& \multicolumn{2}{c}{$\mathbf{E}_\mathbf{SDF}^{\mathcal{N}}$}
		&  \textbf{Timings}\\
		\cmidrule(lr){2-9}
		\cmidrule(lr){10-10}
		\textbf{Method}
		& mean & std.
		& mean & std.
		& mean & std.
		& mean & std. 
		& avg. / shape
		\\
		\midrule
		\textbf{GSD} $(\star)$& \underline{0.0375} & \underline{0.0048} & 0.0673 & 0.0292 
		& \textbf{0.0321} & \underline{0.0060} 
		& 0.0772 & 0.0083 & \underline{11 mins.}\\     
		\textbf{1-Lip}   &  0.0429 & 0.0106 & \textbf{0.0553} & \underline{0.0276} 
		& 0.4806 & 0.1063 
		& 0.0917 & 0.0051 &  74 mins.     \\
		\textbf{HeatSDF}  & 0.2228 & 0.0654 &  0.14313& 0.0366   
		& 0.1431 & 0.0358  
		& \underline{0.0134} & \underline{0.0038} & 32 mins.  \\
		\textbf{Hessian}    & 0.7693 & 0.0628  & 0.6389 & 0.0639  
		& 0.4945 & 0.0842  
		& {0.0283} & 0.0054 & 19 mins.\\
		\textbf{HotSpot}      & 0.0905 & 0.0637 & 0.1415 & 0.0523 
		& 0.4261 & 0.0412 
		& 0.0712 & 0.0082 & \textbf{\hspace{2pt} 6 mins.}\\
		\midrule
		\textbf{Ours}& \textbf{0.0106} & \textbf{0.0041} & \underline{0.0631} & \textbf{0.0123} 
		& \underline{0.0406} & \textbf{0.0056} 
		& \textbf{0.0103} & \textbf{0.0023} & \underline{11 mins.}\\
		\bottomrule
	\end{tabular}
	\caption{Left: SDF evaluation metrics for the SRB dataset. Mean and standard deviation are reported across all shapes. The best results are highlighted in bold and the second-best results are underlined. $(\star)$ denotes a grid based method. Right: Average timings per shape for the different methods.}
	\label{tab:combined}
\end{table*}

\FloatBarrier
\subsubsection{Ablation}
\begin{figure}[htbp!]
	\centering
    \includegraphics[width=0.9\linewidth]{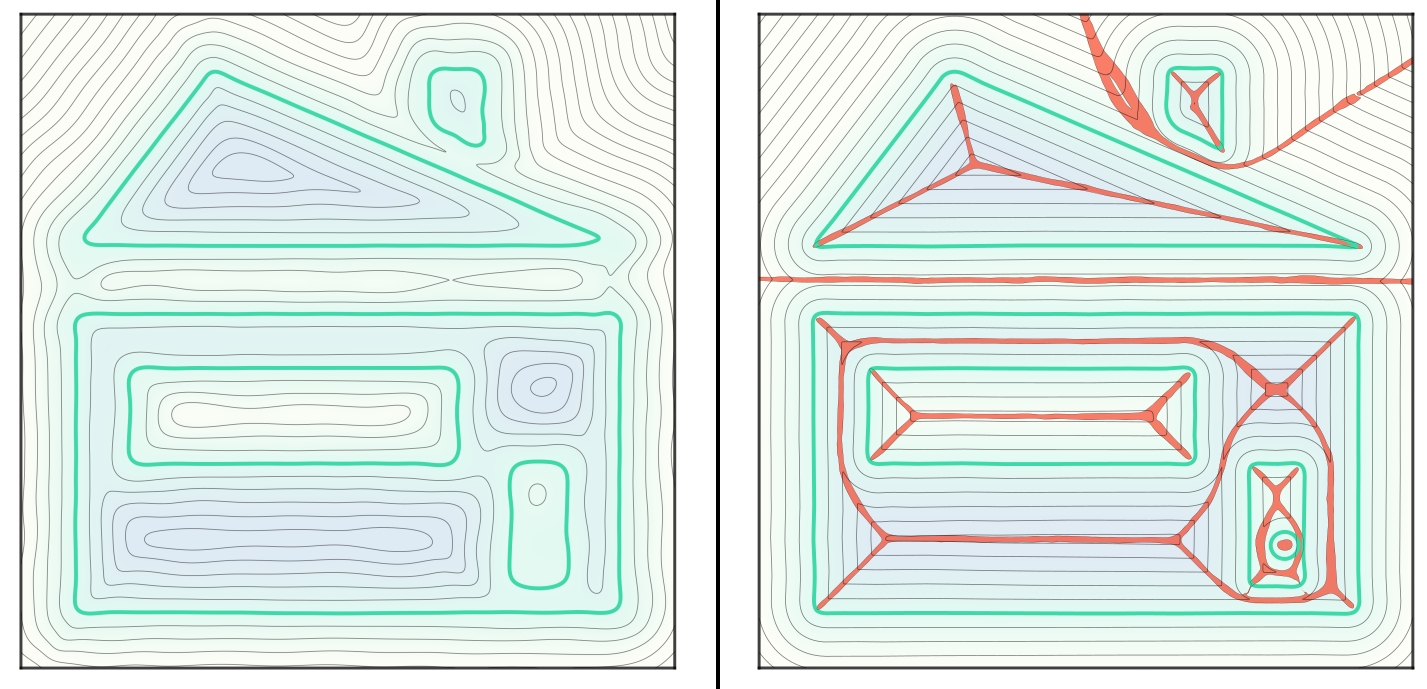}
    \caption{Comparison of our higher order loss with and without phase field on a 2D house geometry. Left: Ours without phase field regularization (i.e. constant $v_\eta \equiv 1$). Right: Ours with phase field (red). While both methods capture the general shape, the level sets in the left figure are significantly less equally spaced and less sharp. Additionally, fine geometric details are lost in the modified version, which can be seen, for example, at the door handle.}
    \label{fig:house_withoutPf}
\end{figure}

The combination of both SDF and phase field in the variational approach in \cref{eq:totalLoss}
is crucial for the performance of our method. In particular, using the eikonal loss in combination with higher order regularization without a phase field 
is not sufficient to recover high quality SDFs (see \cref{fig:house_withoutPf}).\\
\vspace{-0.4cm}
\subsubsection{Sphere Tracing}
\begin{figure}[htbp!]
	\centering
	\begin{tikzpicture}
    \node[anchor=south west, inner sep=0] (img)
        {\includegraphics[width=\linewidth]{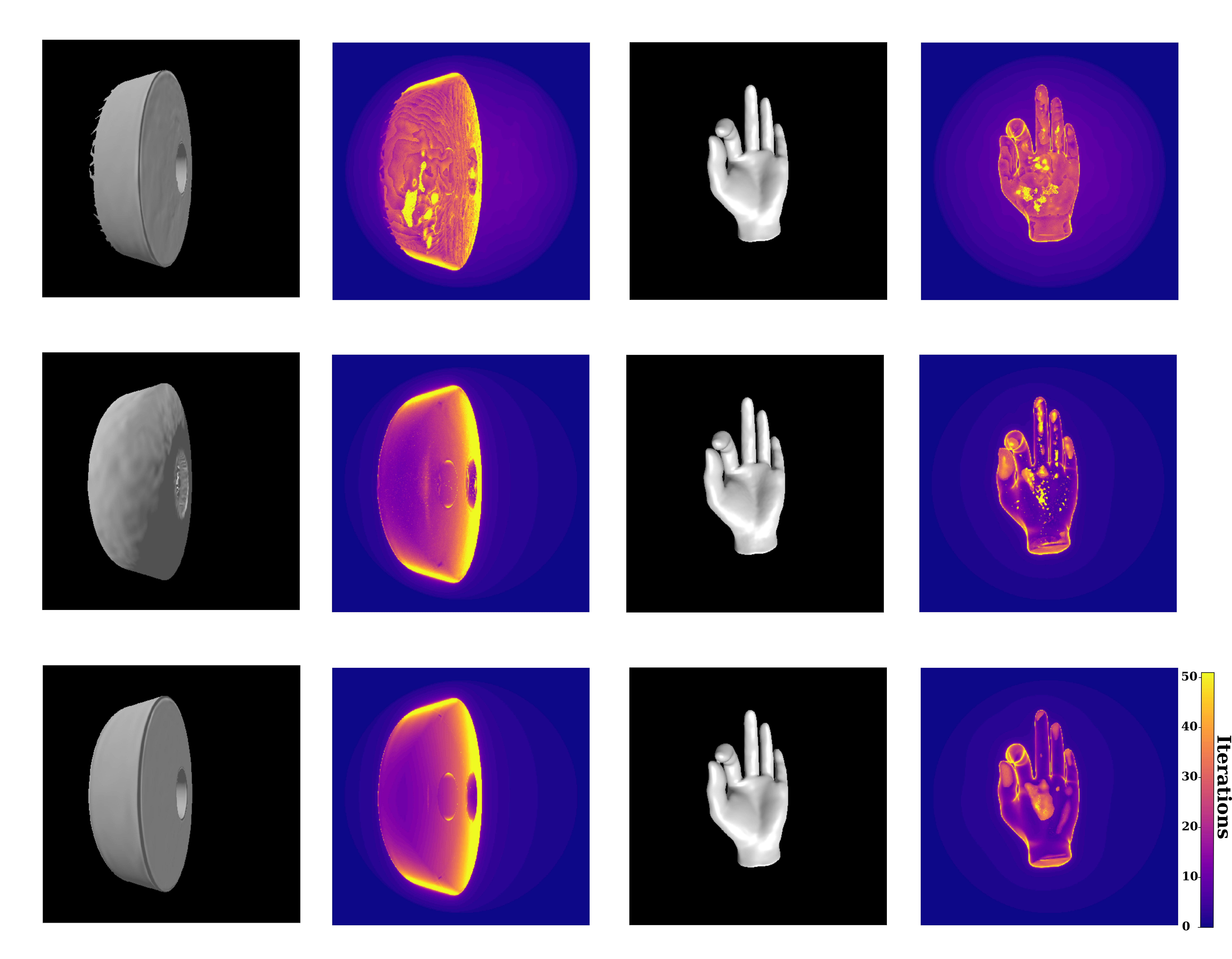}};

    \path (img.south west) rectangle (img.north east);

    \rotatebox{90}{
\node[inner sep=0pt] at (5.4, -0.1) {\textbf{Hessian}};
\node[inner sep=0pt] at (3.2, -0.1) {\textbf{HotSpot}};
\node[inner sep=0pt] at (1.13, -0.1) {\textbf{Ours}};}
\tiny{
\node[inner sep=0pt] at (2.7, 4.41) {Mean: 9.9};
\node[inner sep=0pt] at (2.7, 2.27) {Mean: 6.5};
\node[inner sep=0pt] at (2.7, 0.13) {Mean: 6.2};}
\node[inner sep=0pt] at (6.73, 4.41) {Mean: 5.9};
\node[inner sep=0pt] at (6.73, 2.27) {Mean: 3.6};
\node[inner sep=0pt] at (6.73, 0.13) {Mean: 3.4};
\end{tikzpicture}
\vspace{-0.3cm}
    \caption{Sphere tracing rendering results (in black and white) on a torus with rectangular cross section and a hand shape, for our method (bottom row), HotSpot (middle row), and Hessian (top row). The color plots show the required number of iterations per pixel. These experiments where conducted using the implementation from \cite{HotSpot}.}
    \label{fig:sphere_tracing}
\end{figure}

We demonstrate the practical utility of our SDF method in the context of sphere tracing, see \cite{SphereTracing}.
To this end, we employ the implementation provided by \cite{HotSpot}.
In \cref{fig:sphere_tracing} we show rendering results obtained via sphere tracing, which computes surface intersections along rays using the signed distance values, rather than relying on intermediate mesh extraction methods such as marching cubes.
These results highlight both the computational efficiency and the visual quality of our approach compared to existing methods.

\section{Discussion}
We propose a novel, second-order variational model to compute the SDF of a surface point cloud.
The simultaneously optimized phase field represents the medial axis, i.e. the jump set $\jumpset$ of the 
gradient of the SDF, which promotes an accurate SDF identification. 
The method combines an eikonal loss with a second-order loss term enforcing the characteristic 
linearity of the SDF along gradient directions away from the medial axis,  
and a loss controlling the ($d-1$)-dimensional measure of $\jumpset$.
The simultaneous optimization of the SDF $\phi$ and the phase field $v$ combines
the identification of smooth regions of the SDF 
and the medial axis (the jump set $\jumpset$). This in particular implies a detailed surface reconstruction
and at the same time a reliable SDF in the far field.
Our approach computes SDFs which are close to the groundtruth viscosity solution of the eikonal equation.
Across various experiments we show that our method achieves state-of-the-art surface reconstruction and 
signed distance accuracy, while most other methods focus on either of those. Although our method involves two neural networks, together with second derivatives in the loss functional, we achieve runtimes comparable to those methods 
with only one neural network for the SDF.
We do not require oriented point clouds or a segmentation of inside and outside, because minimizers of the loss 
$\mathcal{L}_\text{total}^\eps$ are already approximations of the signed distance, up to a multiplication with~$-1$.

First-order loss functionals, which involve the non-convex eikonal loss suffer from 
a lack of lower semi-continuity.
In contrast, our second-order variational phase field functional is well-posed, 
in particular the existence of minimizers of the phase field loss functional $\mathcal{L}_\text{total}^\eps$ is ensured.

The phase field approximating the medial axis of the input surface here serves merely as a tool to compute accurate SDFs. We believe however that an accurate reconstruction of the medial axis is possible, following the method proposed in \cite{NeuralSkeleton}, by sampling points on the surface and letting them flow in negative gradient direction, with a step size controlled by the phase field.

\begin{figure}[htbp!]
	\includegraphics[width=\linewidth, trim={0 0.27cm 0 0}, clip]{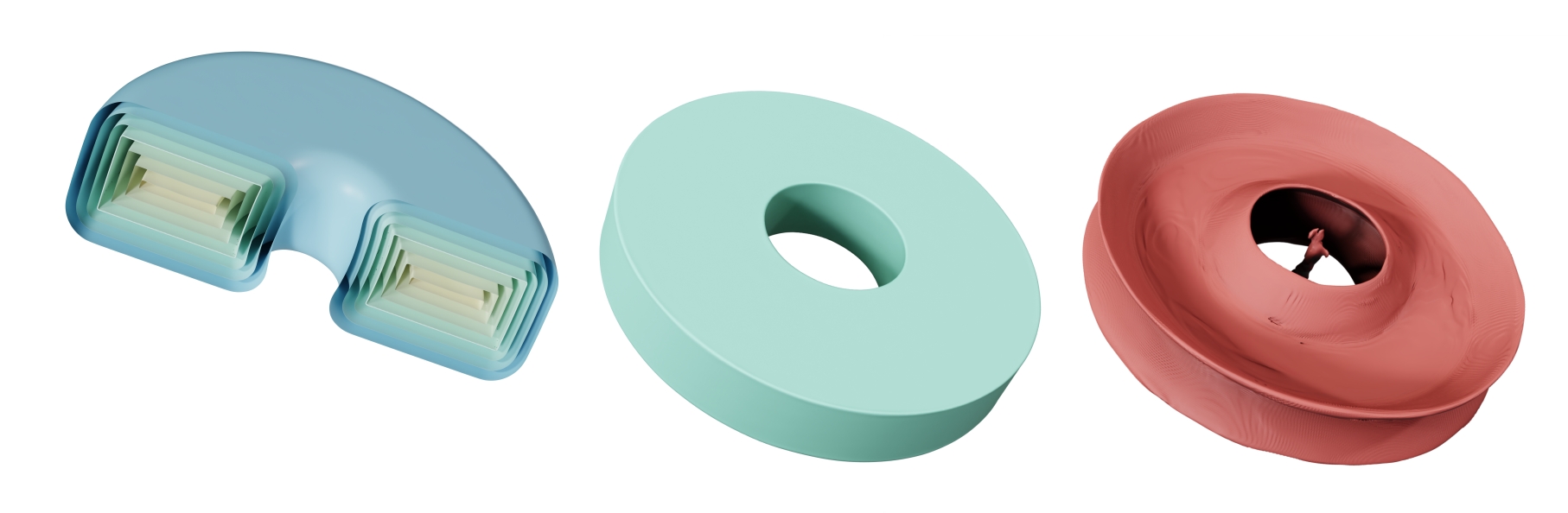}
	\vspace{-0.6cm}
	\caption{Left:  level sets of the neural SDF $\phi_\theta$ of a torus with rectangular cross section. Middle: zero-level set of $\phi_\theta$.
		Right: phase field approximation of $\jumpset$. In addition to the co-dimension 1 interior segments of the jump set, the phase field depicts a segment of the torus's rotational axis, a co-dimension 2 component of $\jumpset$.}
	\label{fig:Ananas_example}
\end{figure}

The behaviour of minimizers for $\eps\to 0$ poses some analytical challenges.
The co-dimension $2$ components of the medial axis are not penalized by the Ambrosio-Tortorelli energy
approximating the $(d-1)$-dimensional measure of $J_{\nabla \phi}$ (cf. \cref{fig:Ananas_example}).

The $\mathcal{L}_\text{exp}$ penalty fosters that minimizers $\phi$ are viscosity solutions, but there is so far no analytical guarantee.

\begin{figure}[htbp!]
	\centering
	 \begin{tikzpicture}
		\node[anchor=south west, inner sep=0](img) {\includegraphics[width=0.8\linewidth, trim={0 0 0 0.2cm}, clip]{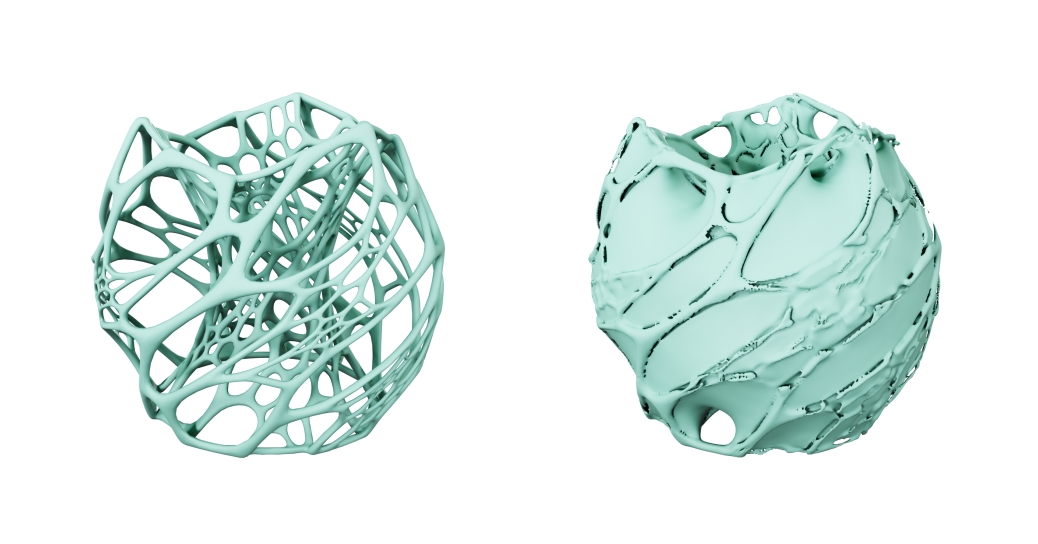}};
		\path (img.south west) rectangle (img.north east);
		\node[inner sep=0pt] at (1.8, 0.2) {\textbf{Ground truth}};
		\node[inner sep=0pt] at (5., 0.2) {\textbf{Ours}};
\end{tikzpicture}
	\vspace{-0.2cm}
	\caption{Example of a shape from the Thingy10k dataset, where our method does not reconstruct the correct SDF.}
	\label{fig:thingiFail}
\end{figure}

Surfaces with very complex topology have an even more complex medial axis. For the fixed set of parameters used in this paper, the resulting phase field does not approximate the medial axis, leading to a severe mismatch in the SDF that also impacts the proper inside-outside identification, cf. \cref{fig:thingiFail}. In fact, our model would require a significantly smaller phase field parameter $\eps$ and larger $\alpha_p$ in $\mathcal{L}_\text{exp}$, and correspondingly an increased network depth and width for an accurate approximation. Currently, this still fails to converge. Here, an adaptation of the optimization strategy is required.

\begin{figure}[htbp!]
	\centering
    \includegraphics[width=0.95\linewidth]{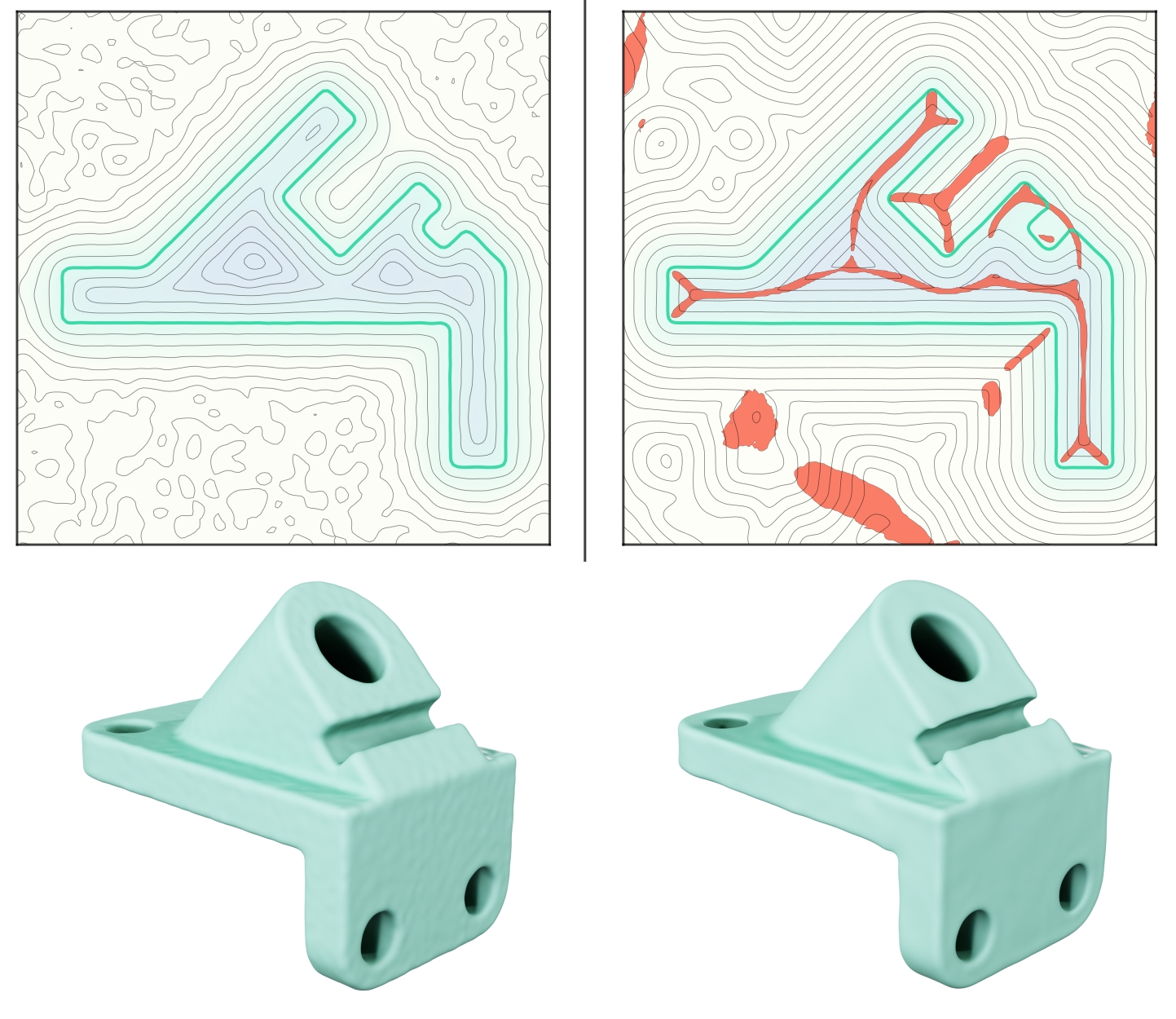}
    \begin{tikzpicture}
\node[inner sep=0pt] at (-4.3, -3.4) {};
\node[inner sep=0pt] at (-2.7, -3.4) {\textbf{HeatSDF}};
\node[inner sep=0pt] at (0.5, -3.4) {\textbf{modified HeatSDF}};
\end{tikzpicture}

	\caption{HeatSDF method (left) and modified HeatSDF method with higher-order loss functional and phase field regularization (right). Top row shows improved level set quality near the surface (right) compared to the original approach (left) for a slice ($y = 0$) of the 3D anchor shape from the SRB data set \cite{SRB_dataset}. The bottom row displays renderings of both results, where the right rendering (modified HeatSDF) exhibits reduced surface noise compared to the left (original HeatSDF).}

	\label{fig:HeatSDF_withPf}
\end{figure}

We believe that our second-order loss, in combination with the Ambrosio–Tortorelli functional, can be integrated into existing SDF methods to improve both surface reconstruction quality and the accuracy of the signed distance approximation. As a proof of concept, we introduce a variation of the two-step first order HeatSDF method, in which the original formulation of the second step loss functional is augmented with an additional higher order phase field regularization term during training, i.e., 
\[
\mathcal{L}_\text{Modified}[\phi, v] := \mathcal{L}_\text{HeatSDF}^\text{2nd}[\phi] + \mathcal{L}_\text{HO}^\eps[\phi,v] + \mathcal{L}_\text{AT}^\eps[v]\,.
\]
This modified approach leads to a sharper and more consistent SDF estimation, compared to the original HeatSDF method
(see \cref{fig:HeatSDF_withPf}). We chose the HeatSDF method for this proof of concept, since it is a first-order method that does not involve solving the Eikonal equation, and thus is inherently different to our approach.

\section*{Acknowlegments}
This work was funded by the Deutsche Forschungsgemeinschaft (DFG, German Research Foundation) – CRC 1720 – 539309657 and via Germany’s Excellence Strategy project 390685813 -- Hausdorff Center for Mathematics. 
\newpage
~\newpage
\begin{spacing}{0.95}
\bibliographystyle{alpha}
\bibliography{sample}

@String{tog = "ACM TOG"}

@article{sharp2019vector,
  title={The vector heat method},
  author={Sharp, Nicholas and Soliman, Yousuf and Crane, Keenan},
  journal={ACM Transactions on Graphics (TOG)},
  volume={38},
  number={3},
  pages={1--19},
  year={2019},
  publisher={ACM New York, NY, USA}
}

@incollection{Bl67,
  citeseer = {http://citeseer.nj.nec.com/context/77000/0},
  author = {Harry Blum},
  booktitle = {Models for the Perception of Speech and Visual Form},
  editor = {Weiant Wathen-Dunn},
  optstatus = {html doi abstract},
  title = {{A} transformation for extracting new descriptors of
           shape},
  address = {Cambridge},
  publisher = {MIT Press},
  year = {1967},
  pages = {362--380},
}

@article{AmDeMa99,
author = {Ambrosio, Luigi and De Lellis, Camillo and Mantegazza, Carlo},
year = {1999},
month = {12},
pages = {327-355},
title = {Line energies for gradient vector fields in the plane},
volume = {9},
journal = {Calculus of Variations and Partial Differential Equations},
doi = {10.1007/s005260050144}
}

@article{JinKohn2000,
  author = {Jin, W. and Kohn, R. V.},
  title = {Singular Perturbation and the Energy of Folds},
  journal = {Journal of Nonlinear Science},
  year = {2000},
  volume = {10},
  number = {3},
  pages = {355--390},
  doi = {10.1007/s003329910014},
  url = {https://doi.org/10.1007/s003329910014}
}

@article{DeLellis02,
     author = {Lellis, Camillo De},
     title = {An example in the gradient theory of phase transitions},
     journal = {ESAIM: Control, Optimisation and Calculus of Variations},
     pages = {285--289},
     year = {2002},
     publisher = {EDP-Sciences},
     volume = {7},
     doi = {10.1051/cocv:2002012},
     mrnumber = {1925030},
     zbl = {1037.49010},
     language = {en},
     url = {https://www.numdam.org/articles/10.1051/cocv:2002012/}
}

@article{AvGi96, 
	title={The distance function and defect energy}, 
	volume={126}, 
	DOI={10.1017/S0308210500023167}, 
	number={5}, 
	journal={Proceedings of the Royal Society of Edinburgh: Section A Mathematics}, 
	author={Aviles, Patricio and Giga, Yoshikazu}, 
	year={1996}, 
	pages={923–938}
}

@article{AvGi87,
  author = {Aviles, Patricio and Giga, Yoshikazu},
  title = {A mathematical problem related to the physical theory of liquid crystal configurations},
  journal = {Proc. Centre Math. Anal. Austral. Nat. Univ.},
  volume = {12},
  year = {1987},
  pages = {1--16},
  url = {https://maths.anu.edu.au/files/CMAProcVol12-AvilesGiga_2.pdf}
}

@article{MoMo77,
    title = {Un Esempio Di {$\Gamma$}-Convergenza},
    author = {Modica, Luciano and Mortola, Stefano},
    year = {1977},
    journal = {Boll. Un. Mat. Ital. B (5)},
    volume = {14},
    number = {1},
    pages = {285--299}
}

@ARTICLE{MuSh89,
	author = {Mumford, David and Shah, Jayant},
	title = {Optimal approximations by piecewise smooth functions and associated
	variational problems},
	journal = {Comm. Pure Appl. Math.},
	year = {1989},
	volume = {42},
	pages = {577--685},
	number = {5},
	doi = {10.1002/cpa.3160420503},
	printed = {1}
}

@ARTICLE{AmTo92,
  author = {Luigi Ambrosio and Vincenzo M. Tortorelli},
  title = {On the approximation of free discontinuity problems},
  journal = {Bollettino dell{}'Unione Matematica Italiana, Sezione B},
  year = {1992},
  volume = {6},
  pages = {105--123},
  number = {7}
}

@article{SRB_dataset,
author = {Berger, Matthew and Levine, Joshua A. and Nonato, Luis Gustavo and Taubin, Gabriel and Silva, Claudio T.},
title = {A benchmark for surface reconstruction},
year = {2013},
issue_date = {April 2013},
publisher = {Association for Computing Machinery},
address = {New York, NY, USA},
volume = {32},
number = {2},
issn = {0730-0301},
doi = {10.1145/2451236.2451246},
journal = {ACM Trans. Graph.},
month = apr,
articleno = {20},
numpages = {17}
}

@article{ResNet,
  title={Deep Residual Learning for Image Recognition},
  author={Kaiming He and X. Zhang and Shaoqing Ren and Jian Sun},
  journal={2016 IEEE Conference on Computer Vision and Pattern Recognition (CVPR)},
  year={2015},
  pages={770-778},
}

@misc{SIREN,
      title={Implicit Neural Representations with Periodic Activation Functions}, 
      author={Vincent Sitzmann and Julien N. P. Martel and Alexander W. Bergman and David B. Lindell and Gordon Wetzstein},
      year={2020},
      eprint={2006.09661},
      archivePrefix={arXiv},
      primaryClass={cs.CV},
      url={https://arxiv.org/abs/2006.09661}, 
}

@inproceedings{HotSpot,
      title={HotSpot: Signed Distance Function Optimization with an Asymptotically Sufficient Condition},
      author={Wang, Zimo and Wang, Cheng and Yoshino, Taiki  and Tao, Sirui and Fu, Ziyang  and Li, Tzu-Mao},
      year={2025},
      booktitle={CVPR},
    }

@article{GSD,
author = {Feng, Nicole and Crane, Keenan},
title = {A Heat Method for Generalized Signed Distance},
year = {2024},
issue_date = {July 2024},
publisher = {Association for Computing Machinery},
address = {New York, NY, USA},
volume = {43},
number = {4},
issn = {0730-0301},
url = {https://doi.org/10.1145/3658220},
doi = {10.1145/3658220},
journal = {ACM Trans. Graph.},
month = jul,
articleno = {92},
numpages = {19}
}

@inproceedings{Li21Phase,
  title={Phase Transitions, Distance Functions, and Implicit Neural Representations},
  author={Yaron Lipman},
  booktitle={International Conference on Machine Learning},
  year={2021},
  url={https://api.semanticscholar.org/CorpusID:235435938}
}

@article{MaMoMi24,
title = {Phase-field modeling of fracture with physics-informed deep learning},
journal = {Computer Methods in Applied Mechanics and Engineering},
volume = {429},
pages = {117104},
year = {2024},
issn = {0045-7825},
doi = {https://doi.org/10.1016/j.cma.2024.117104},
url = {https://www.sciencedirect.com/science/article/pii/S0045782524003608},
author = {M. Manav and R. Molinaro and S. Mishra and L. {De Lorenzis}},
}

@article{FrMa98,
title = {Revisiting brittle fracture as an energy minimization problem},
journal = {Journal of the Mechanics and Physics of Solids},
volume = {46},
number = {8},
pages = {1319-1342},
year = {1998},
issn = {0022-5096},
doi = {https://doi.org/10.1016/S0022-5096(98)00034-9},
url = {https://www.sciencedirect.com/science/article/pii/S0022509698000349},
author = {G.A. Francfort and J.-J. Marigo},
}

@article{BoFrMa00,
title = {Numerical experiments in revisited brittle fracture},
journal = {Journal of the Mechanics and Physics of Solids},
volume = {48},
number = {4},
pages = {797-826},
year = {2000},
issn = {0022-5096},
doi = {https://doi.org/10.1016/S0022-5096(99)00028-9},
url = {https://www.sciencedirect.com/science/article/pii/S0022509699000289},
author = {B. Bourdin and G.A. Francfort and J-J. Marigo},
}

@article{HeatSDF26,
author = {Weidemaier, Samuel and Hartwig, Florine and Sassen, Josua and Conti, Sergio and Ben-Chen, Mirela and Rumpf, Martin},
title = {SDFs from Unoriented Point Clouds using Neural Variational Heat Distances},
journal = {Computer Graphics Forum},
volume = {n/a},
number = {n/a},
pages = {e70296},
doi = {https://doi.org/10.1111/cgf.70296},
url = {https://onlinelibrary.wiley.com/doi/abs/10.1111/cgf.70296},
eprint = {https://onlinelibrary.wiley.com/doi/pdf/10.1111/cgf.70296},
year = {2026},
}

@article{crane2017heat,
  title={The heat method for distance computation},
  author={Crane, Keenan and Weischedel, Clarisse and Wardetzky, Max},
  journal={Communications of the ACM},
  volume={60},
  number={11},
  pages={90--99},
  year={2017},
  publisher={ACM New York, NY, USA}
}

@inproceedings{yang2023steik,
author = {Yang, Huizong and Sun, Yuxin and Sundaramoorthi, Ganesh and Yezzi, Anthony},
title = {StEik: stabilizing the optimization of neural signed distance functions and finer shape representation},
year = {2023},
publisher = {Curran Associates Inc.},
address = {Red Hook, USA},
booktitle = {Proceedings of the 37th International Conference on Neural Information Processing Systems},
articleno = {617},
numpages = {12},
location = {New Orleans, LA, USA},
series = {NIPS '23}
}

@inproceedings{ben2022digs,
  title={Digs: Divergence guided shape implicit neural representation for unoriented point clouds},
  author={Ben-Shabat, Yizhak and Koneputugodage, Chamin Hewa and Gould, Stephen},
  booktitle={Proceedings of the IEEE/CVF Conference on Computer Vision and Pattern Recognition},
  address={New Orleans, USA},
  pages={19323--19332},
  year={2022}
}

@article{wang2023neural,
  title={Neural-singular-{Hessian}: Implicit neural representation of unoriented point clouds by enforcing singular {Hessian}},
  author={Wang, Zixiong and Zhang, Yunxiao and Xu, Rui and Zhang, Fan and Wang, Peng-Shuai and Chen, Shuangmin and Xin, Shiqing and Wang, Wenping and Tu, Changhe},
  journal={ACM Transactions on Graphics (TOG)},
  volume={42},
  number={6},
  pages={1--14},
  year={2023},
  publisher={ACM New York, NY, USA}
}

@inproceedings{hessian2,
 author = {Wang, Ruian and Wang, Zixiong and Zhang, Yunxiao and Chen, Shuangmin and Xin, Shiqing and Tu, Changhe and Wang, Wenping},
 booktitle = {Advances in Neural Information Processing Systems},
 editor = {A. Oh and T. Naumann and A. Globerson and K. Saenko and M. Hardt and S. Levine},
 pages = {63515--63528},
 publisher = {Curran Associates, Inc.},
 title = {Aligning Gradient and Hessian for Neural Signed Distance Function},
 url = {https://proceedings.neurips.cc/paper_files/paper/2023/file/c87bd5843849884e9430f1693b018d71-Paper-Conference.pdf},
 volume = {36},
 year = {2023}
}

@article{essakine2025where,
title={Where Do We Stand with Implicit Neural Representations? A Technical and Performance Survey},
author={Amer Essakine and Yanqi Cheng and Chun-Wun Cheng and Lipei Zhang and Zhongying Deng and Lei Zhu and Carola-Bibiane Sch{\"o}nlieb and Angelica I Aviles-Rivero},
journal={Transactions on Machine Learning Research},
issn={2835-8856},
year={2025},
aaurl={https://openreview.net/forum?id=QTsJXSvAI2},
note={Survey Certification},
publisher={Journal of Machine Learning Research Inc.}
}

@inproceedings{marschner2023constructive,
  title={Constructive solid geometry on neural signed distance fields},
  author={Marschner, Zo{\"e} and Sell{\'a}n, Silvia and Liu, Hsueh-Ti Derek and Jacobson, Alec},
  booktitle={SIGGRAPH Asia 2023 conference papers},
  address={Sydney, Australia},
  pages={1--12},
  year={2023}
}

@article{liu2024real,
  title={Real-time collision detection between general {SDFs}},
  author={Liu, Pengfei and Zhang, Yuqing and Wang, He and Yip, Milo K and Liu, Elvis S and Jin, Xiaogang},
  journal={Computer Aided Geometric Design},
  volume={111},
  pages={102305},
  year={2024},
  publisher={Elsevier}
}

@inproceedings{mehta2022level,
  title={A level set theory for neural implicit evolution under explicit flows},
  author={Mehta, Ishit and Chandraker, Manmohan and Ramamoorthi, Ravi},
  booktitle={European Conference on Computer Vision},
  address={Tel Aviv, Israel},
  pages={711--729},
  year={2022},
  organization={Springer}
}

@inproceedings{coiffier20241,
  title={{1-Lipschitz} Neural Distance Fields},
  author={Coiffier, Guillaume and B{\'e}thune, Louis},
  booktitle={Computer Graphics Forum},
  volume={43},
  number={5},
  pages={e15128},
  year={2024},
  organization={Wiley Online Library}
}

@article{crane2013geodesics,
  title={Geodesics in heat: A new approach to computing distance based on heat flow},
  author={Crane, Keenan and Weischedel, Clarisse and Wardetzky, Max},
  journal={ACM Transactions on Graphics (TOG)},
  volume={32},
  number={5},
  pages={1--11},
  year={2013},
  publisher={ACM New York, NY, USA}
}

@article{yeh2010template,
	title={Template-based 3d model fitting using dual-domain relaxation},
	author={Yeh, I-Cheng and Lin, Chao-Hung and Sorkine, Olga and Lee, Tong-Yee},
	journal={IEEE Transactions on Visualization and Computer Graphics},
	volume={17},
	number={8},
	pages={1178--1190},
	year={2010},
	publisher={IEEE}
}

@article{Thingi10K,
  title={Thingi10K: A Dataset of 10,000 3D-Printing Models},
  author={Zhou, Qingnan and Jacobson, Alec},
  journal={Preprint arXiv:1605.04797},
  year={2016}
}

@article{Ki34,
author = {Kirszbraun, Mojżesz D.},
journal = {Fundamenta Mathematicae},
language = {ger},
number = {1},
pages = {77-108},
title = {Über die {Z}usammenziehende und {L}ipschitzsche {T}ransformationen},
url = {http://eudml.org/doc/212681},
volume = {22},
year = {1934},
}

@article{Kingma2014AdamAM,
	title={Adam: A Method for Stochastic Optimization},
	author={Diederik P. Kingma and Jimmy Ba},
	journal={CoRR},
	year={2014},
	volume={abs/1412.6980},
	aaurl={https://api.semanticscholar.org/CorpusID:6628106}
}

@book{Fe69,
  title={Geometric Measure Theory},
  author={Federer, Herbert},
  isbn={978-3-642-62010-2},
  url={https://link.springer.com/book/10.1007/978-3-642-62010-2},
  year={1969},
  publisher={Springer-Verlag Berlin Heidelberg},
  series={Die Grundlehren der mathematischen Wissenschaften},
  volume={153}
}

@BOOK{Ev98,
  title = {Partial Differential Equations},
  publisher = {American Mathematical Society},
  year = {1998},
  author = {Evans, L.~C.}
}

@inbook{PyTorch,
author = {Paszke, Adam and Gross, Sam and Massa, Francisco and Lerer, Adam and Bradbury, James and Chanan, Gregory and Killeen, Trevor and Lin, Zeming and Gimelshein, Natalia and Antiga, Luca and Desmaison, Alban and K\"{o}pf, Andreas and Yang, Edward and DeVito, Zach and Raison, Martin and Tejani, Alykhan and Chilamkurthy, Sasank and Steiner, Benoit and Fang, Lu and Bai, Junjie and Chintala, Soumith},
title = {PyTorch: an imperative style, high-performance deep learning library},
year = {2019},
publisher = {Curran Associates Inc.},
address = {Red Hook, NY, USA},
booktitle = {Proceedings of the 33rd International Conference on Neural Information Processing Systems},
articleno = {721},
numpages = {12}
}

@article{QuaNets,
title = {Universal approximation with quadratic deep networks},
journal = {Neural Networks},
volume = {124},
pages = {383-392},
year = {2020},
issn = {0893-6080},
doi = {https://doi.org/10.1016/j.neunet.2020.01.007},
url = {https://www.sciencedirect.com/science/article/pii/S0893608020300095},
author = {Fenglei Fan and Jinjun Xiong and Ge Wang},
}

@article{SphereTracing,
  title={Sphere tracing: a geometric method for the antialiased ray tracing of implicit surfaces},
  author={John C. Hart},
  journal={The Visual Computer},
  year={1996},
  volume={12},
  pages={527-545},
}

@misc{MeshToSDF,
	author = {Kleineberg, Marian},
	title = {Mesh-to-sdf: Calculate signed distance fields for arbitrary meshes},
	publisher = {GitHub},
	year = {2021},
	url = {https://github.com/marian42/mesh_to_sdf},
}

@article{NeuralSkeleton,
  title   = {Neural Skeleton: Implicit Neural Representation Away from the Surface},
  author  = {Matt{\'e}o Cl{\'e}mot and Julie Digne},
  journal = {Computers \& Graphics},
  volume  = {114},
  pages   = {368--378},
  year    = {2023},
  doi     = {10.1016/j.cag.2023.06.012},
}

@inproceedings{PointSkeleton,
  title     = {Predicting Animation Skeletons for 3D Articulated Models via Volumetric Nets},
  author    = {Zhan Xu and Yang Zhou and Evangelos Kalogerakis and Karan Singh},
  booktitle = {Proceedings of the International Conference on 3D Vision (3DV)},
  pages     = {298--307},
  year      = {2019},
}

@article{ImplicitSkeleton,
  title   = {Deep Medial Fields},
  author  = {Daniel Rebain and Ke Li and Vincent Sitzmann and Soroosh Yazdani and Kwang Moo Yi and Andrea Tagliasacchi},
  journal = {arXiv preprint arXiv:2106.03804},
  year    = {2021},
}
\end{spacing}
\appendix
\section{Proof of \cref{thm:ExistenceOfMinimizers}}
\label{sec:appproofs}
Let $(\phi_i^\eps,v_i^\eps)_{i\in\N}\subset H^2(\Omega)\times H^1(\Omega)$ be a minimizing sequence, i.e. 
 \[ \inf_{(\phi,v)}\mathcal{L}_\text{total}^\eps[\phi,v] =  
 \lim\limits_{i\rightarrow \infty} \mathcal{L}_\text{total}^\eps[\phi_i^\eps, v_i^\eps]
\leq C < \infty\,.\]
Here, $C>0$ is a universal constant that may change it's value from line to line.
Using this boundedness and the Poincaré-type estimate 
\[\Vert \phi \Vert^2_{L^2(\Omega)} \leq C ( \Vert \nabla \phi \Vert^2_{L^2(\Omega)} + \Vert \phi \Vert^2_{L^2(\surface)})\] 
there exists a constant $C$, s.t.
\[\|\phi_i^\eps\|_{H^2(\Omega)} + \|v_i^\eps\|_{H^1(\Omega)} \leq C \quad \forall i\in \N\,.\]
By the reflexivity of the spaces $H^2(\Omega)$ and $H^1(\Omega)$, there exist 
$(\phi^\eps,v^\eps)\in H^2(\Omega)\times H^1(\Omega)$, such that $\phi_i^\eps \rightharpoonup\phi^\eps$ weakly in $H^2(\Omega)$ and $v_i^\eps\rightharpoonup v^\eps$ weakly in $H^1(\Omega)$ for a subsequence (not relabeled). By Rellich's theorem, there exist further subsequences, such that $\phi_i^\eps \rightarrow \phi^\eps $ strongly in $H^1(\Omega)$, and $v_i^\eps\rightarrow v^\eps$ strongly in $L^2(\Omega)$. By the weak lower semi-continuity of the norm, and by dominated convergence, we hence have 
\[\liminf_{i\rightarrow\infty}\mathcal{L}_\text{AT}^\eps[v_i^\eps] \geq \mathcal{L}_\text{AT}^\eps[v^\eps]\,,\]
and  
\begin{align*}
&\lim_{i\rightarrow\infty}\left(\mathcal{L}_\text{eik}^\eps[\phi_i^\eps] + \mathcal{L}_\text{exp}[\phi_i^\eps] + {\mathcal{L}_\text{recon}^\eps}[\phi_i^\eps]\right) \\
&= \mathcal{L}_\text{eik}^\eps[\phi^\eps] + \mathcal{L}_\text{exp}[\phi^\eps] + {\mathcal{L}_\text{recon}^\eps}[\phi^\eps]\,.
\end{align*}
Again, by weak lower semi-continuity of the norm, we have 
\[\liminf_{i\rightarrow\infty} \int_\Omega\eps^2|D^2\phi_i^\eps|^2 \d x \geq \int_\Omega\eps^2|D^2\phi^\eps|^2 \d x\,.
\]
Using again the boundedness of the total loss, 
the sequence $(v_i^\eps D^2\phi_i^\eps\nabla \phi_i^\eps)_{i\in\N}$ is bounded in $L^2(\Omega;\R^d)$, 
hence there exists $\Theta^\eps\in L^2(\Omega;\R^d)$, such that 
$$v_i^\eps D^2\phi_i^\eps\nabla \phi_i^\eps \rightharpoonup \Theta^\eps \quad\text{weakly in }L^2(\Omega;\R^d)$$ 
for a subsequence. On the other hand, $(v_i^\eps)_{i\in\N}$ and $(\nabla\phi_i^\eps)_{i\in\N}$ are converging strongly in 
$L^6(\Omega)$ and $L^6(\Omega;\R^d)$ for $d\in \{2,3\}$, see \cite{Ev98}. Hence, 
$$v_i^\eps D^2\phi_i^\eps\nabla \phi_i^\eps\rightharpoonup v^\eps D^2\phi^\eps\nabla\phi^\eps\quad \text{weakly in }L^{\frac{6}{5}}(\Omega;\R^d)\,.$$ This implies $\Theta^\eps = v^\eps D^2\phi^\eps\nabla\phi^\eps$ almost everywhere and by the weak lower semicontinuity of the norm 
\[\liminf_{i\rightarrow\infty}\mathcal{L}_\text{HO}^\eps[\phi_i^\eps,v_i^\eps] \geq \mathcal{L}_\text{HO}^\eps[\phi^\eps,v^\eps]\,.\] 
Finally, we obtain 
\[\inf_{(\phi,v)}\mathcal{L}_\text{total}^\eps[\phi,v] =\liminf_{i\rightarrow\infty}\mathcal{L}_\text{total}^\eps[\phi_i^\eps,v_i^\eps] \geq \mathcal{L}_\text{total}^\eps[\phi^\eps,v^\eps]\,,\] 
which implies that $(\phi^\eps,v^\eps)$ is a minimizer of the total loss $\mathcal{L}_\text{total}^\eps$.

\section{Proof of \cref{thm:limsup}}
\label{sec:RecoverySequence}

	We can extend $\phi$ to all of $\R^d$, having Lipschitz constant $1$, cf.~\cite{Ki34}.
	We construct the recovery sequence $(\phi_\eps,v_\eps)_\eps$ in the following way: 
	As in \cite{AmTo92}, we introduce a decreasing positive sequence $b_\eps$ with $\tfrac{b_\eps}{\eps} \rightarrow 0$ and 
	$\tfrac{\eps^2}{b_\eps} \rightarrow 0$ and define 
	\begin{align}
\label{eq:vEpsConstruction}
v^\eps(x) \coloneqq \begin{cases}
0 \quad &\text{in }(\jumpset)_{b_\eps}, \\
1 - \exp\left(\frac{b_\eps - \mathrm{dist}(x,\jumpset)}{2\eps}\right)&\text{else.}
\end{cases}.
\end{align}
Under the above assumptions, following \cite{AmTo92}, the convergence
	\begin{align*}
		\lim_{\eps\rightarrow 0} \mathcal{L}_\text{AT}^\eps[v^\eps] &= \mathcal{H}^{d-1}(\jumpset)\,.
	\end{align*}
	is ensured.
	In order to construct $\phi_\eps$, let $\vartheta_{b_\eps}$ be a smooth cutoff-function, such that 
	\[ 
	\vartheta_{b_\eps} = 0 \;\; \text{on }\R^d\setminus(\jumpset)_{b_\eps}\,,\quad\vartheta_{b_\eps}= 1\;\;\text{on }(\jumpset)_{\frac{b_\eps}{2}}\,,\quad \|\nabla\vartheta_{b_\eps}\| \leq \frac{4}{b_\eps}\,. 
	\]
	Furthermore, we take into account the mollifier $\rho^{b_\eps}$ defined as
	\begin{align*}
	& \rho^{b_\eps}(x) \coloneqq b_\eps^{-d} \rho(\frac{x}{b_\eps})\\
	\text{where }&\rho(x)\coloneqq \begin{cases}
	C_d \exp(\frac{1}{\|x\|^2 - 1})\quad&\text{, if }\|x\|< 1 \\
	0\quad&\text{, else}
	\end{cases}\,.
\end{align*}
We define the recovery sequence $\phi^\eps$ as 
	\begin{align}
	\label{eq:recSeq}
	\phi^\eps \coloneqq \phi + \left( (\rho^{b_\eps}\ast\phi) - \phi \right)\vartheta_{b_\eps}\,,
	\end{align}
	i.e. $\phi^\eps\in W^{2,2}(\Omega)$ and $\phi^\eps = \phi$ on $\Omega\setminus(\jumpset)_{b_\eps}$.

\begin{figure}[htbp!]
\centering
\def\svgwidth{0.75\linewidth}
\normalsize
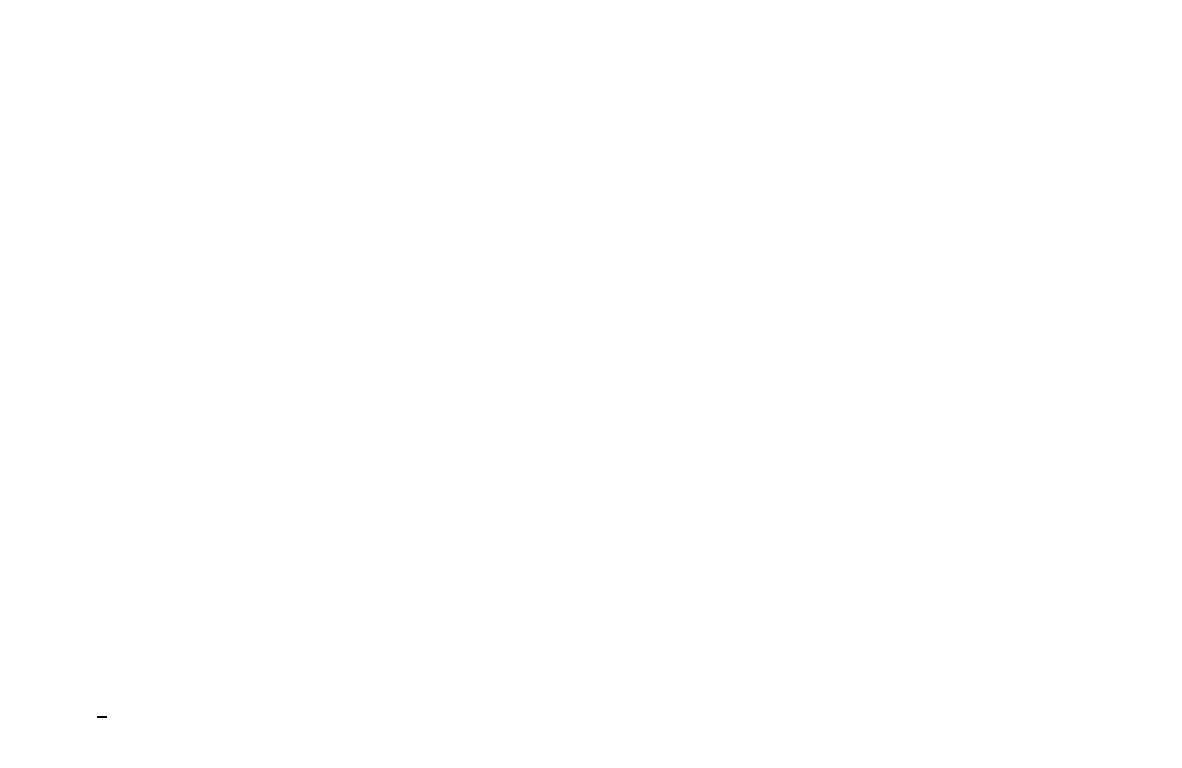
\caption{On a slice in direction normal to $\jumpset$ (red) the constructed $\phi^\eps$ and $v^\eps$ are sketched for a 
given solution $\phi$ of the eikonal equation with $\phi=0$ on $\surface$ (in green).}
\label{fig:RecoveryConstructionSketch}
\end{figure}

A sketch of $v^\eps$ and $\phi^\eps$ is displayed in \cref{fig:RecoveryConstructionSketch}. 

	By the $1$-Lipschitz continuity of $\phi$, we have the pointwise estimate
	\begin{align}
	&|(\rho^{b^\eps}\ast\phi)(x) - \phi(x)| = \left| \int_{B_{b_\eps}(0)}(\phi(x-y)-\phi(x))\rho^{b^\eps}(y)\d y \right|
	\nonumber\\ \leq& \sup_{z\in B_{b_\eps}(x)}|\phi(z) - \phi(x)|\leq \|\nabla\phi\|_\infty b_\eps = b_\eps\,.
	\label{eq:pointwiseError}
	\end{align}
	The definition of $\phi^\eps$ ensures in the limit for $\eps\to 0$
	\begin{align*}
		\lim_{\eps\rightarrow 0} \mathcal{L}_\text{exp}[\phi^\eps] &= \mathcal{L}_\text{exp}[\phi]\,,
	\end{align*}
	by dominated convergence, and also the limit
	\begin{align*}
		\lim_{\eps\rightarrow 0} \mathcal{L}_\text{recon}^\eps[\phi^\eps] = 0\,,	
	\end{align*}
	using \cref{eq:pointwiseError}, and the fact that $\phi=0$ on $\surface$.
	The gradient of $\phi^\eps$ reads as 
	\begin{align}
	\label{eq:recSeqGrad}
	\nabla \phi^\eps = \nabla \phi +  \left((\rho^{b_\eps}\ast\nabla\phi)-\nabla\phi\right)\vartheta_{b_\eps} + \left((\rho^{b_\eps}\ast\phi)-\phi\right) \nabla\vartheta_{b_\eps}\,.
	\end{align}
	With this we can estimate
	\begin{align*}
	\left|\|\nabla\phi^\eps\|^2 - 1\right| \leq& |\vartheta_{b_\eps}|^2\|(\rho^{b_\eps}\ast\nabla\phi) - \nabla\phi\|^2 \\
	&+ \|\nabla\vartheta_{b_\eps}\|^2 |(\rho^{b_\eps}\ast\phi) - \phi|^2\\ 
	&+ 2 \|\nabla\phi\|\;|\vartheta_{b_\eps}|\;\|(\rho^{b_\eps}\ast\nabla\phi)- \nabla\phi\|\\
	&+ 2 \|\nabla\phi\|\;|(\rho^{b_\eps}\ast\phi)-\phi|\;\|\nabla\vartheta_{b_\eps}\|\\ 
	&+ 2 |\vartheta_{b_\eps}|
   \;\|(\rho^{b_\eps}\ast\nabla\phi)-\nabla\phi\|\\
&\qquad\cdot
   \|\nabla\vartheta_{b_\eps}\|
   \;|(\rho^{b_\eps}\ast\phi)-\phi|\\
	\leq& 4 + \frac{16}{b_\eps^2}b_\eps^2 + 4 +  \frac{8}{b_\eps} b_\eps +  \frac{16}{b_\eps} b_\eps = 48\,,
	\end{align*}
	were we used  $\|(\rho^{b_\eps}\ast\nabla\phi) - \nabla\phi\|\leq 2$ and $|\vartheta_{b_\eps}|\leq~1$. 
	Hence, the eikonal-part of the loss can be estimated as 
	\begin{align*}
		\lim_{\eps\rightarrow 0} \mathcal{L}_\text{eik}^\eps[\phi^\eps] &= \lim_{\eps\rightarrow 0}\left( \frac1\eps \int_\Omega\left|\|\nabla\phi^\eps\|^2 - 1\right|^2\d x \right)\\
		&= \lim_{\eps\rightarrow 0} \left(\frac1\eps \int_{(\jumpset)_{b_\eps}}\left|\|\nabla\phi^\eps\|^2 - 1\right|^2\d x\right)\\
		&\leq \lim_{\eps\rightarrow 0}\left( \frac{48^2|(\jumpset)_{b_\eps}|}{\eps}\right)\\ &= \lim_{\eps\rightarrow 0}\left(48^2 \frac{|(\jumpset)_{b_\eps}|}{2b_\eps}\frac{2b_\eps}{\eps}\right) = 0\,.
	\end{align*}

	It is left to show that 
	\begin{align*}
		\mathcal{L}_\text{HO}^\eps[\phi^\eps,v^\eps] =& \int_\Omega (v_\eps^2 + \eps^2)\|D^2\phi^\eps\nabla\phi^\eps\|^2 \d x\\ = &\int_{\Omega\setminus (\jumpset)_{b_\eps}} (v_\eps^2 + \eps^2)\|D^2\phi\nabla\phi\|^2 \d x \\&+ \int_{(\jumpset)_{b_\eps}} \eps^2\|D^2\phi^\eps\nabla\phi^\eps\|^2 \d x\\ =& \int_{(\jumpset)_{b_\eps}} \eps^2\|D^2\phi^\eps\nabla\phi^\eps\|^2 \d x
	\end{align*}
	vanishes in the limit as $\eps\rightarrow 0$. Here we used that $D^2\phi\nabla\phi = 0$ on ${\Omega\setminus\jumpset}$. We split the integral on the right hand side into two integrals over the sets $(\jumpset)_{\tfrac{b_\eps}{2}}$ and $(\jumpset)_{b_\eps} \setminus (\jumpset)_{\tfrac{b_\eps}{2}}$, respectively.
	On $(\jumpset)_{\tfrac{b_\eps}{2}}$ we can estimate
	\[
	\|D^2\phi^\eps\| = \|\nabla \rho^{b_\eps}\ast \nabla \phi_\eps\| \leq C \frac1{b_\eps}\,.
	\]
	On $(\jumpset)_{b_\eps} \setminus (\jumpset)_{\tfrac{b_\eps}{2}}$, we can write the Hessian as
	\vspace{-0.1cm}
	\begin{align*}
		D^2\phi^\eps =& D^2\phi + \left((\rho^{b_\eps}\ast D^2\phi) - D^2\phi\right)\vartheta_{b_\eps}\\ &+  2\left((\rho^{b_\eps}\ast \nabla\phi) - \nabla\phi\right)\otimes \nabla\vartheta_{b_\eps}\\ &+ \left((\rho^{b_\eps}\ast \phi) - \phi\right)D^2\vartheta_{b_\eps}\,.
	\end{align*}
\vspace{-0.1cm}
	{Here, the first two terms are bounded in $L^2(\Omega\setminus \jumpset)$, the third term can be bounded pointwise by  
	$\tfrac{16}{b_\eps}$, and for the fourth term is bounded pointwise by $C\tfrac{b_\eps}{b_\eps^2}$. This implies}
	\begin{align*}
		\lim_{\eps\rightarrow 0}\mathcal{L}_\text{HO}^\eps[\phi^\eps,v^\eps]&=\lim_{\eps\rightarrow 0} \int_{(\jumpset)_{b_\eps}} \eps^2\|D^2\phi^\eps\nabla\phi^\eps\|^2 \d x\\ &\leq \lim_{\eps\rightarrow 0} C\left( \eps^2 + \frac{\eps^2}{b_\eps} \right) = 0\,,
	\end{align*}
	where we again used $\|\nabla \phi^\eps\| \leq C$ and $${\lim\limits_{\eps\rightarrow 0}\tfrac{|(\jumpset)_{b_\eps}|}{2b_\eps} = \mathcal{H}^{d-1}(\jumpset) < \infty}\,.$$
\vspace{-1.0cm}
\section{2D Evaluation}
\label{Appendix:2D}
\begin{figure}[htbp!]
	\centering
	 {\includegraphics[width=0.95\linewidth]{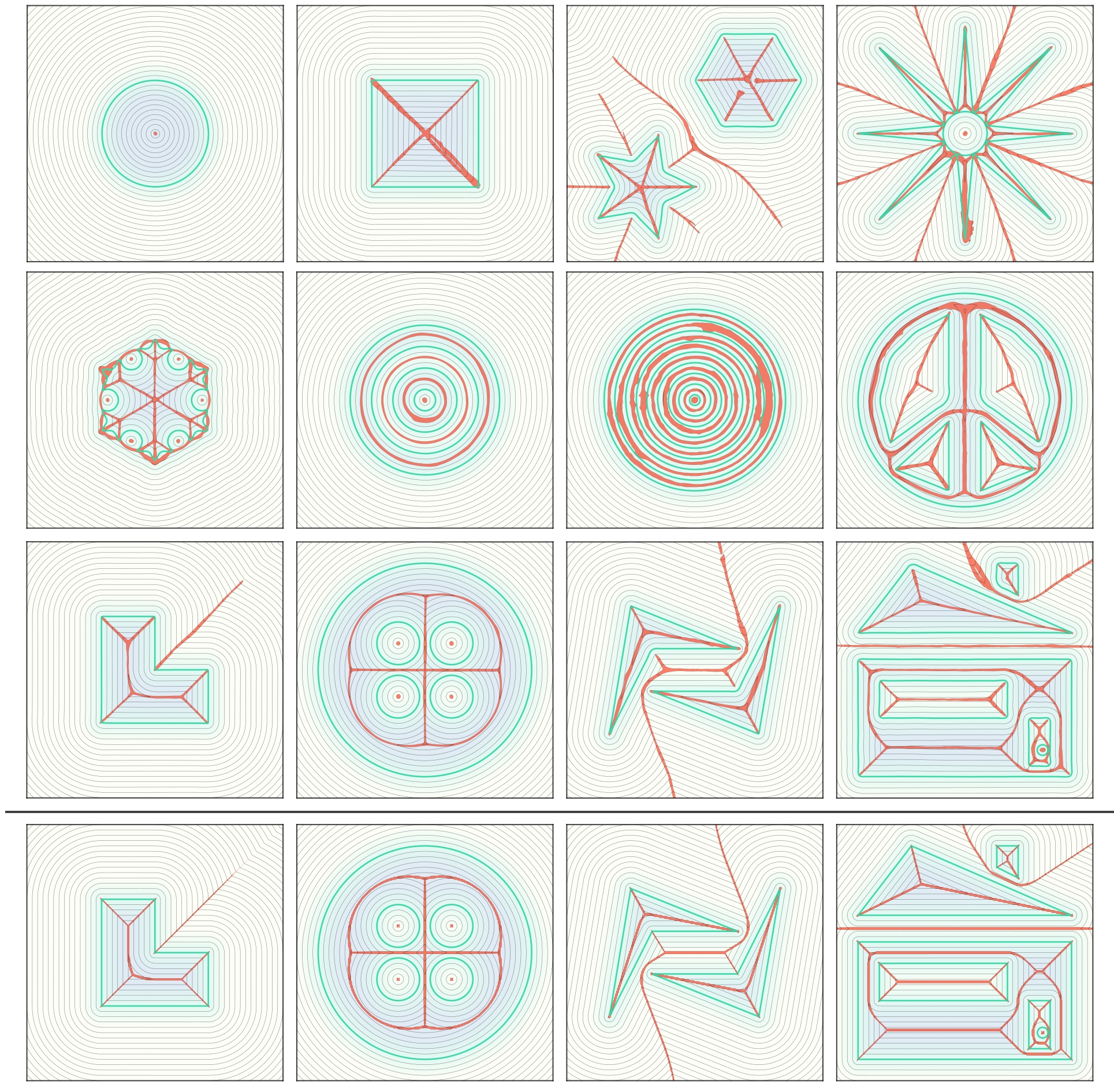}}
    \begin{tikzpicture}
        \centering
    \tiny{
    \rotatebox{90}{
        \node[font=\bfseries\small] at (0, 1.6) { };
        \node[font=\bfseries\tiny] at (2.74, 0.2) {Ground truth};
        \node[font=\bfseries\tiny] at (6.7, 0.2) {Ours (SDF \& PF)};
    }}
\end{tikzpicture}
\vspace{-2.0cm}\caption{First three rows: Computed solutions for the 2D-dataset of \cite{HotSpot}. Zero-level set of SDF in green, $0.25$-sublevel set of phase field in red. Bottom row: Ground truth solution. Zero-level set of SDF in green, jump set $\jumpset$ in red. }
\label{fig:2dExperiments}
\end{figure}

\vspace{-0.2cm}
For the experiments in 2D, we start the training with the weights
\vspace{-0.2cm}
\begin{align*}
(\gamma_\text{HO}, \gamma_\text{AT}, \gamma_\mathrm{recon}, \gamma_\mathrm{eik}, \gamma_\text{exp})_\text{Phase 1} = (10,0.2,10,0.1,100)\,.
\end{align*}
Following the schedule described in \cref{sec:Scheduling}, we end with the final weights
\vspace{-0.2cm}
\begin{align*}
	(\gamma_\text{HO}, \gamma_\text{AT}, \gamma_\mathrm{recon}, \gamma_\mathrm{eik}, \gamma_\text{exp})_\text{Final} = (10,0.2,10,0.1,1)\,.
\end{align*}
In \cref{fig:2dExperiments}, we provide additional qualitative results of the $2d$-dataset from \cite{HotSpot}, together with the phase field approximation of the medial axis. 
\vspace{-0.2cm}

\vspace{-0.4cm}
\FloatBarrier
\section{3D Evaluation}
\label{Appendix:3D}
\begin{figure*}[htbp!]
  {\includegraphics[width=\linewidth]{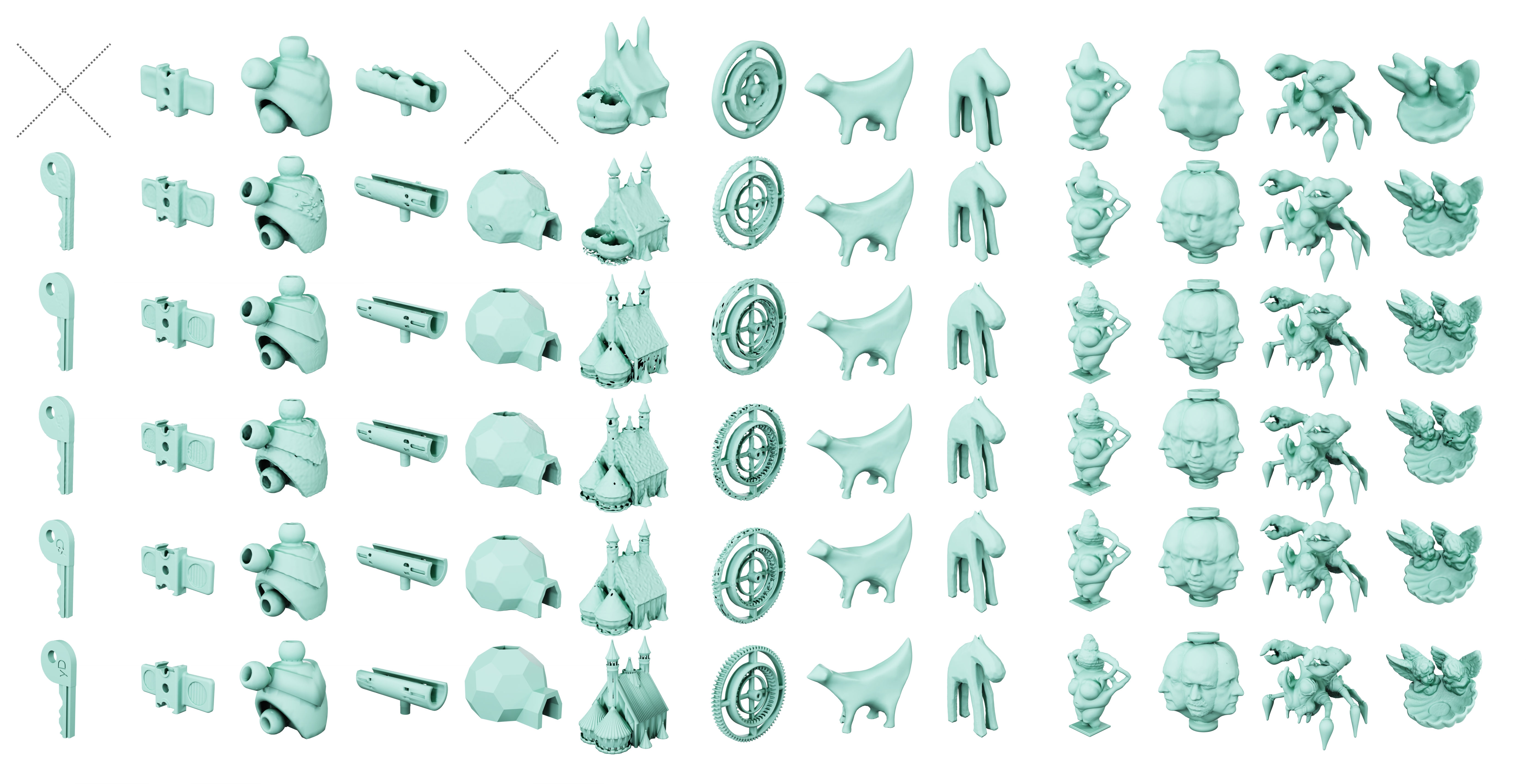}}
    \begin{tikzpicture}
        \centering
    \tiny{
    \rotatebox{90}{
        \node[font=\bfseries\small] at (0, 1.6) { };
        \node[font=\bfseries\tiny] at (2.78, 0.05) {Ground truth};
        \node[font=\bfseries\tiny] at (4.2, 0.05) {Ours};
        \node[font=\bfseries\tiny] at (5.6, 0.05) {HotSpot};
        \node[font=\bfseries\tiny] at (7.1, 0.05) {Hessian};
        \node[font=\bfseries\tiny] at (8.45, 0.05) {HeatSDF};
        \node[font=\bfseries\tiny] at (9.7, 0.05) {1-Lip};
    }}
\end{tikzpicture}
\vspace{-2.0cm}  
\caption{Marching Cubes results for the zero-level sets of the computed SDFs for a selection of shapes from the Thingy10k dataset. The first seven shapes are from those without tag specification. The last six shapes are with the tag 'sculpture' or 'scan'. Runs that did not converge to an extractable geometry have been marked with an "X". }
\label{fig:thingy10k_vis}
\end{figure*}

For the experiments in 3D, we start the training with the weights
\begin{align*}
(\gamma_\text{HO}, \gamma_\text{AT}, \gamma_\mathrm{recon}, \gamma_\mathrm{eik}, \gamma_\text{exp})_\text{Phase 1} = (1,0.02,0.01,0.05,500)\,.
\end{align*}
Following the schedule described in \cref{sec:Scheduling}, we end with the final weights
\begin{align*}
(\gamma_\text{HO}, \gamma_\text{AT}, \gamma_\mathrm{recon}, \gamma_\mathrm{eik}, \gamma_\text{exp})_\text{Final} = (2.5,0.2,0.5,0.2,200)\,.
\end{align*}
We provide additional experiments on a subset of the Thingi10k data set \cite{Thingi10K}. In \cref{fig:thingy10k_vis}, a qualitative comparison of the zero-level set surface reconstruction on a selection of shapes is shown for different neural SDF methods. 

For each shape, we extract 10k randomly sampled surface points for the training.
For the evaluation we compare with the neural methods (Hessian, HotSpot, 1-Lip and HeatSDF). 
In \cref{tab:thingy10k_results}, quantitative results evaluated on the whole subset of shapes are displayed. 1-Lip \cite{coiffier20241} and HeatSDF \cite{HeatSDF26} did not converge on some of these shapes. We have excluded these shapes in the error evaluation of these methods.

\begin{table*}[htbp!]
\centering

\setlength{\tabcolsep}{2.8pt}
\begin{tabular}{
    l
    cc
    cc
    cc
    cc
    cc
    cc
    cc
}
\toprule
& \multicolumn{2}{c}{$\mathbf{d_{C}}$}
& \multicolumn{2}{c}{$\mathbf{d_{H}}$}
& \multicolumn{2}{c}{$\mathbf{E_{n}}$}
& \multicolumn{2}{c}{$\mathbf{E}_\mathbf{eik}^\Omega$}
& \multicolumn{2}{c}{$\mathbf{E}_\mathbf{eik}^\mathcal{N}$}
& \multicolumn{2}{c}{$\mathbf{E}_\mathbf{SDF}^{\Omega}$}
& \multicolumn{2}{c}{$\mathbf{E}_\mathbf{SDF}^{\mathcal{N}}$}\\
\cmidrule(lr){2-15}

\textbf{Method}
& mean & std.
& mean & std.
& mean & std.
& mean & std.
& mean & std.
& mean & std.
& mean & std.
 \\
\midrule  
\textbf{1-Lip} & 0.0536 & 0.0146 & 0.1576 & 0.0938 & 0.1413 & 0.1115 & \textbf{0.0728} & \underline{0.0326} & \underline{0.1109} & \underline{0.0602} & 0.8607 & 0.1761 & 0.1109 & 0.0602\\ 
\textbf{HeatSDF}& 0.0448 & 0.0208 & 0.1689 & 0.1206 & 0.0893 & \textbf{0.0947}  &  0.2136 & \textbf{0.0209} & 0.1588 & 0.0724 &\underline{0.3215} &0.0942&\textbf{0.0172}&0.0103 \\
\textbf{Hessian} & \textbf{0.0110} & \textbf{0.0043} & \underline{0.0472} & \underline{0.0744} & \underline{0.0753} & 0.1106 & 0.7615 & 0.0598 & 0.8185 & 0.2042 &
  0.4841 & 0.1089 &
 \underline{0.0318} & \underline{0.0090} \\
\textbf{HotSpot}   & 0.0130 & 0.0089 &\textbf{0.0566} & \textbf{0.0665} & 0.0818 &{0.1049}     & 0.1069 & 0.0742 & 0.2143 & 0.0774
 & 0.4266 & \underline{0.0613} 
&{0.0671} & \textbf{0.0010}\\
\midrule
\textbf{Ours} 
&  \underline{0.0124} & \underline{0.0073}& {0.0847} & 0.1125 & \textbf{0.0741} &\underline{0.1024}& \underline{0.0769} & 0.0829
& \textbf{0.0189} & \textbf{0.0164}
 & \textbf{0.0253} & \textbf{0.0079}  & 0.0871 & 0.0174 \\
\bottomrule
\end{tabular}

\caption{Error analysis for the Thingy10k dataset. Mean and standard deviation are reported across all shapes. The best results are highlighted in bold and the second-best results are underlined.}
\label{tab:thingy10k_results}
\end{table*}

The experiments are conducted on a subset of models from the Thingi10K dataset. 
Specifically, we consider the following 40 models, identified by their Thing IDs:
\{1368069, 521602, 52073, 87522, 49420, 399567, 95778, 42025, 100679, 153957, 
131602, 63497, 200688, 134622, 225950, 591249, 1452679, 384085, 814665, 370886, 
353684, 64194, 84990, 61258, 73986, 133077, 133582, 91684, 68382, 131438, 
65942, 896329, 84997, 73075, 101902, 103538, 354371, 39087, 55031, 133079\}.
Additionally, we used the model with Thing ID 53159, for our teaser figure. The hand geometry used for the spheretracing experiment is from \cite{yeh2010template}.\\
The point cloud for the box (\cref{fig:Box_example}) and the torus with rectangular cross section (\cref{fig:Ananas_example}) were created by ourselbes.

\end{document}